\documentclass[10pt,twocolumn,letterpaper]{article}
\usepackage{titling}
\usepackage{3dv}

\usepackage{times}
\usepackage{epsfig}
\usepackage{graphicx}
\usepackage{amsmath}
\usepackage{amssymb}
\usepackage{booktabs}
\usepackage{caption}
\usepackage{subcaption}

\usepackage[pagebackref=true,breaklinks=true,letterpaper=true,colorlinks,bookmarks=false]{hyperref}

\DeclareMathOperator*{\argmin}{arg\,min} %JED

\threedvfinalcopy % *** Uncomment this line for the final submission
\begin{document}

%%%%%%%%% TITLE
\title{3D Point Cloud Registration with Multi-Scale Architecture 
\\ and Unsupervised Transfer Learning}

\author{Sofiane Horache
\and Jean-Emmanuel Deschaud 
\and François Goulette\\
MINES ParisTech, PSL University, Centre for Robotics, 75006 Paris, France\\
\small{firstname.surname@mines-paristech.fr}\\
}
\maketitle
%\thispagestyle{empty}

%%%%%%%%% ABSTRACT
\begin{abstract}
  We propose a method for generalizing deep learning for 3D point cloud registration on new, totally different datasets. It is based on two components, MS-SVConv and UDGE.
  Using Multi-Scale Sparse Voxel Convolution, MS-SVConv is a fast deep neural network that outputs the descriptors from point clouds for 3D registration between two scenes. 
  UDGE is an algorithm for transferring deep networks on unknown datasets in a unsupervised way.
  The  interest  of  the  proposed  method  appears  while  using the two components, MS-SVConv and UDGE, together as a whole, which leads to state-of-the-art results on real world registration datasets such as 3DMatch, ETH and TUM. The code is publicly available at \url{https://github.com/humanpose1/MS-SVConv}.
\end{abstract}

%%%%%%%%% BODY TEXT
\section{Introduction}

%Point cloud registration has become important in many applications, such as SLAM, pattern recognition and 3D reconstruction~\cite{zeng20163dmatch, 7299195}. Registration can be used online (for example in LiDAR SLAM pipeline for loop closure detection) or offline (for example for 3D reconstruction of indoor scenes). is a key component in mapping environments, maps used for localization and perception in robotics systems. \JED{Papier de référence}.

With the increasing number of 3D sensors and 3D data production, the task of consolidating overlapping point clouds, which is called registration, has become a major issue in many applications. Registration can be used online (e.g., in a LiDAR SLAM pipeline for loop closure detection) or offline (e.g., in 3D reconstructions of RGB-D indoor scenes~\cite{7299195} or for outdoor LiDAR map building for autonomous vehicles).

However, point cloud registration can be challenging with real-world scans because of noise, incomplete 3D scan, outliers, and so on. There are numerous existing approaches to this issue; however, recently, deep learning approaches have become very popular, especially for learning descriptors and for computing transformations. 
%These data-driven approaches have been very effective for solving registration problems~\cite{zeng20163dmatch, deng2018ppffoldnet, gojcic2018perfect, gojcic2020learning, aoki2019pointnetlk, bai2020d3feat, deng2018ppfnet,choy2019fully,choy2020deep}, especially by learning on large datasets with the ground truth pose as supervision.
These data-driven approaches, especially those that involve learning on large datasets with the ground truth pose as supervision, have been very effective at solving registration problems~\cite{zeng20163dmatch, deng2018ppffoldnet, gojcic2018perfect, gojcic2020learning, aoki2019pointnetlk, bai2020d3feat, deng2018ppfnet,choy2019fully,choy2020deep}.

\begin{figure}[t]
    \centering
    \includegraphics[width=\linewidth]{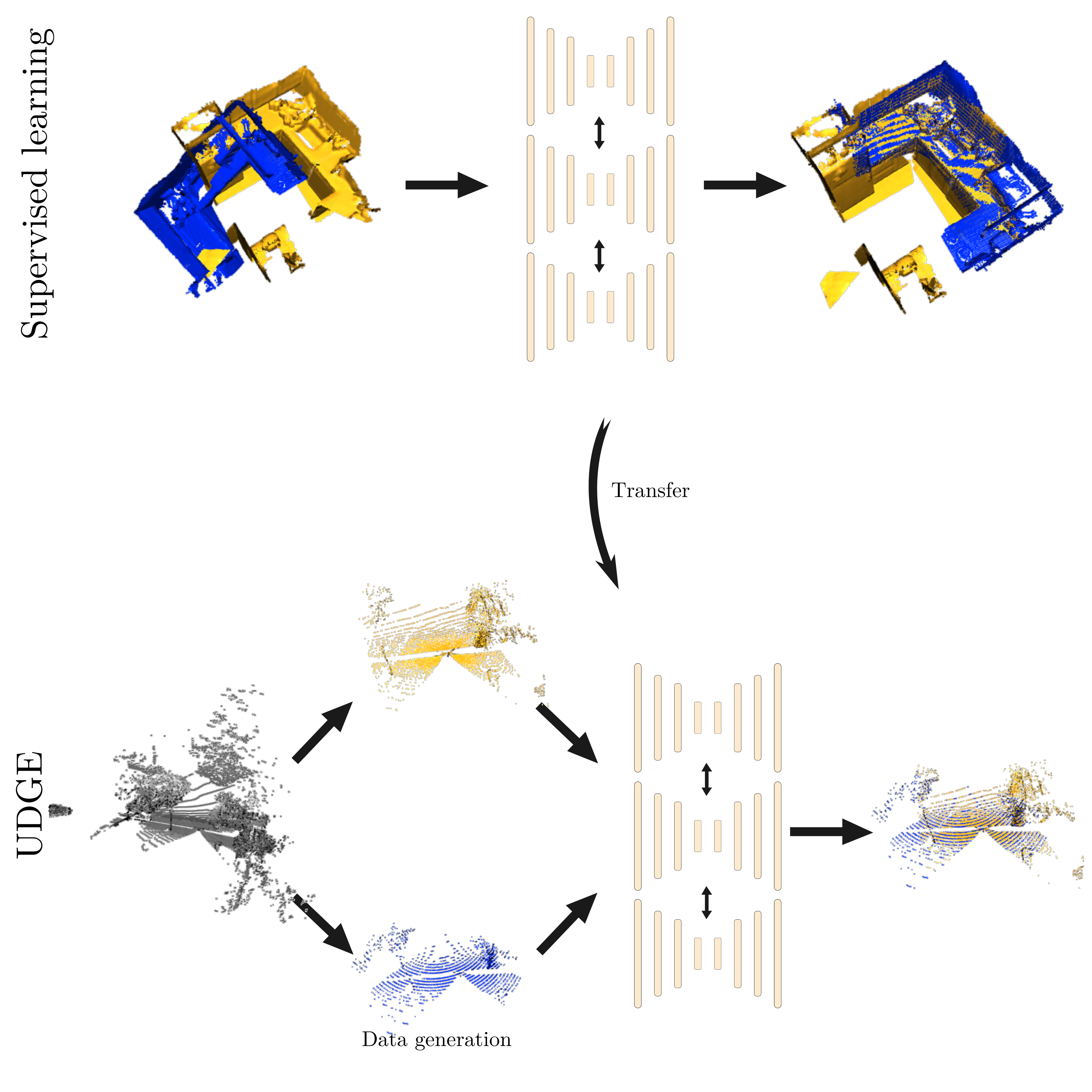}
    \caption{Summary of the proposed method. We can use MS-SVConv for supervised training on a huge dataset, and then we can transfer it in an unsupervised fashion on a smaller dataset.}
    \label{fig:catcheye}
\end{figure}

However, acquiring the ground truth pose can be very costly, and many applications have only small datasets. This is why recent methods are focused on improving the generalization capability of the networks. Therefore, we propose Multi-scale Sparse
Voxel Convolution (MS-SVConv), a U-Net-based method to tackle the problem of registration (see Figure~\ref{fig:catcheye} for a summary of the proposed method). 
Using sparse convolutions (such as MinkowskiNet from Choy \etal~\cite{choy2019fully} or SPVNAS from Tang \etal~\cite{tang2020searching}), MS-SVConv is fast and efficient.
However, contrary to other U-Net methods, such as FCGF~\cite{choy2019fully}, MS-SVConv can compute meaningful descriptors that can generalize with new datasets. Moreover, we also propose Unsupervised transfer learning with Data GEneration (UDGE), so the network can generalize on unknown and small datasets.
With this new transfer learning strategy, we show that we can pre-train MS-SVConv on a synthetic dataset, and transfer it to a totally different real-world dataset.

%By using multi-scale U-Net, our method can easily handle density variations, hence, it can adapt to a dataset with a different density. Thus, by using self supervised fine-tuning, our method can generalize to unknown and small datasets without ground truth poses. unsupervised transfer learning means that contrary to supervised method (such as FCGF \cite{choy2019fully} or D3feat \cite{bai2020d3feat}) there is no supervision, the ground truth pose is not necessary to learn these features. Even with a small dataset, we can learn a meaningful representation to registrate scene.
%Our method allow keeping best of both worlds by being flexible, fast and effective like U-Net based method but also allow generalizing across numerous datasets like Patch-Based methods.\\

In other words, our contributions are the following:
\begin{itemize}
    \item  We propose a method for generalizing deep learning for 3D point cloud registration on new, totally different datasets, using transfer learning. It is based on two components, MS-SVConv which is a shared multi-scale architecture, and UDGE which is an unsupervised transfer learning using data generation.
    \item We evaluate the interest of the proposed method on real-world datasets such as 3DMatch, ETH, and TUM datasets. We show that we can pre-train MS-SVConv on a synthetic dataset, and transfer it to a totally different real-world dataset.
\end{itemize}
MS-SVConv and UDGE, used together as a whole, show the interest of the proposed method. 
%-------------------------------------------------------------------------

\section{Related works}
Many methods have been proposed to tackle rigid point cloud registration. One of the most well-known algorithms is the Iterative Closest Point (ICP) algorithm~\cite{121791} due to its simplicity, modularity, and effectiveness~\cite{rusinkiewicz_efficient_2001, gelfand_geometrically_2003, bouaziz_sparse_2013}. Many subsequent variants of the ICP algorithm have been proposed, (e.g.,~\cite{rusinkiewicz_efficient_2001, bouaziz_sparse_2013}). However, one of the main drawbacks of this algorithm is that it does not work for all initial transformations (the convergence of ICP is local). In addition, the algorithm does not work in every case (when the overlap between two point clouds is too small, or when there are many outliers). Some methods have been proposed to solve the global registration problem (4PCS~\cite{aiger_4-points_nodate}, GoICP~\cite{yang_go-icp:_2016}).
On the other hand, researchers have developed many methods to compute handcrafted descriptors, such as Spin Image~\cite{johnson_using_1999}, SHOT~\cite{Salti2014SHOTUS}, or FPFH~\cite{rusu_fast_2009} on 3D point clouds. 
Although Hana \etal ~\cite{hana_comprehensive_2018} completed a global review of these descriptors, they do not work well for real-world scans, as excessive noise or occlusion decreases the descriptors' adaptability.
\subsection{Deep learning on 3D point clouds}
Unlike images, deep learning on 3D point clouds is challenging because point clouds are unstructured data. Numerous methods have been suggested, such as Multi-View CNN~\cite{su2015multiview} and 3DCNN~\cite{maturana2015}, but these methods require a huge amount of memory. %A large revolution in deep learning in point cloud has been PointNet~\cite{qi2016pointnet} and PointNet++~\cite{qi2017pointnet}, which have allowed many applications~\cite{deng2018ppffoldnet, aoki2019pointnetlk, deng2018ppfnet, Zhao_2019_CVPR}. 
PointNet ~\cite{qi2016pointnet} and Point-Net++  ~\cite{qi2017pointnet} catalyzed a revolution in deep learning in point clouds and have made possible many new applications ~\cite{deng2018ppffoldnet, aoki2019pointnetlk, deng2018ppfnet, Zhao_2019_CVPR}.
% CIte application of pointnet
Recently, many works have tried to generalize convolution for sparse data, such as KPConv~\cite{thomas2019KPConv}, Minkowski~\cite{choy20194d}, and RandLaNet~\cite{hu2019randla}. Guo \etal \cite{guo2020survey} conducted a comprehensive survey of deep learning on 3D point clouds.

\subsection{Deep learning for point cloud registration}
Recently, registration methods based on deep learning have shown remarkable results on synthetic datasets, such as ModelNet~\cite{Wu_2015_CVPR}, and RGB-D datasets, such as 3DMatch~\cite{zeng20163dmatch}. 
%In deep learning methods for point cloud registration, we can observe three different trends:
We observe three trends for these methods:
\begin{itemize}
    \item The end-to-end methods compute the transformation directly thanks to a fully differentiable framework (such as DCP~\cite{wang2019deep} or PointnetLK~\cite{aoki2019pointnetlk}).
    
    \item Deep descriptors are computed locally on patches (such as 3DSmoothNet~\cite{gojcic2018perfect} or  DIP~\cite{Poiesi2021}). The transformation is achieved by matching the descriptors.
    
    \item Deep descriptors are computed simultaneously on every point through use of an architecture similar to U-Net (such as FCGF~\cite{choy2019fully} or D3Feat~\cite{bai2020d3feat}). The transformation is also achieved by matching descriptors.
    
\end{itemize}

\paragraph{End-to-end methods:}
Many end-to-end methods have been proposed to deal with the problem of registration, such as PointNetLK~\cite{aoki2019pointnetlk}, deep closest point~\cite{wang2019deep}, DeepVCP~\cite{lu2019deepvcp}, RPMnet~\cite{yew2020rpmnet}, and PRNet~\cite{Wang_2019_NeurIPS}. Some end-to-end methods are also unsupervised~\cite{Wang_2019_NeurIPS, huang2020featuremetric}.  However, most end-to-end methods were tested only on ModelNet, which is a synthetic dataset, and not a real dataset. Actually, Huang \etal~\cite{huang2020featuremetric} tested their method on 7-Scene~\cite{shotton20137scene}, and the results are promising. However, they have not evaluated their method on 3DMatch dataset. 
%However, they have not shown comparison with other deep learning methods than PointnetLK~\cite{aoki2019pointnetlk} and 3DSmoothNet~\cite{gojcic2018perfect}. 
Deep Global Registration~\cite{choy2020deep} is an end-to-end method that performs very well on a real dataset and is built on FCGF~\cite{choy2019fully} but did not demonstrate an ability to generalize efficiently. According to Choy \etal~\cite{choy2020deep}, Deep Closest Point~\cite{wang2019deep}, and PointNetLK~\cite{aoki2019pointnetlk} do not work on datasets such as 3DMatch.
Recently, PointNetLK has been revisited~\cite{Li_2021_CVPR} and has shown promising generalization results on 3DMatch.

\paragraph{Patch-based descriptor matching methods:}
Since 3DMatch~\cite{zeng20163dmatch} was introduced, many improvements have been made to the patch-based descriptor matching method. PPFNet~\cite{deng2018ppfnet} uses a PointNet architecture to compute local descriptors, and with global pooling, can compute a global descriptor to bring the global context of the scene. PPF Foldnet~\cite{deng2018ppffoldnet} and Capsule Net~\cite{Zhao_2019_CVPR} use an autoencoder to compute descriptors without supervision. With 3DSmoothNet~\cite{gojcic2018perfect}, great improvements have been made on the 3DMatch dataset~\cite{zeng20163dmatch}, and 3DSmoothNet showed great generalization capabilities on the ETH dataset~\cite{Pomerleau:2012}. 3DSmoothNet first computes the local reference frame to orient the patches and then, uses a 3D convolutional neural network on them with a smooth density to compute descriptors. Following the success of 3DSmoothNet, MultiView~\cite{Li_2020_CVPR}, DIP~\cite{Poiesi2021}, GeDI~\cite{Poiesi2021gedi}, and SpinNet~\cite{ao2020SpinNet} also improved upon the generalization capabilities from the 3DMatch dataset~\cite{zeng20163dmatch} to the ETH dataset~\cite{Pomerleau:2012}.
But even if these methods are rotation invariant, patch-based methods are very slow in inference.
Therefore, in real-world conditions, patch-based methods are inoperable, because of patch extraction.
Moreover, because each patch is treated individually, their design is usually not flexible enough to add new capabilities (keypoint detector, end-to-end extension, feature pre-training for semantic segmentation). 
%But even if these methods are rotation invariant, patch-based methods are very slow. Moreover, their design is not flexible enough to add new capabilities (keypoint detector, end-to-end extension, feature pre-training for semantic segmentation) because usually, each patch is treated individually. 

\paragraph{U-Net based descriptor matching method:}
FCGF~\cite{choy2019fully} is one of the first U-Net methods that was applied to point cloud registration; thanks to the architecture and the sparse voxel convolutions, this method performs very well. It is also much faster than patch-based methods in inference. D3Feat~\cite{bai2020d3feat} uses Kernel Point convolutions~\cite{thomas2019KPConv} and jointly computes the detector and a descriptor.
%By using sparse convolutions, FCGF~\cite{choy20194d} stays faster, more effective and more memory efficient than D3Feat. 
FCGF and D3Feat show poor generalization capabilities.
PREDATOR~\cite{huang2020predator} is composed of a Graph Neural Network and a cross attention module to compute meaningful descriptors, even if the overlap between two scenes is low.
These works reveal that U-Net-based methods are more flexible than patch-based methods. They can be coupled easily with a detector or with an end-to-end method~\cite{choy2020deep}, and can also be used in multi-scene registration methods~\cite{gojcic2020learning}. The main problem with these methods is that they have poor generalization capabilities on unknown datasets. 
% The proposed method is mainly inspired from FCGF but has much better generalization capabilities while keeping its advantages (speed, efficiency, modularity).
The proposed method, which is mainly inspired by FCGF, maintains its advantages (speed, efficiency, modularity) while showing much better generalization capabilities than FCGF.

\section{Proposed method}
%In this section, we will present the proposed method and we will show how we can generalize on different dataset thanks to pre-training on a source dataset.

%In this section, we will first describe the problem of registration and remind the principle of U-Net based methods. Then, we will explain the proposed contribution to improve the U-Net by using a multi-scale architecture. We will finally explain how we fine-tune our network on unknown scenes in self-supervised way.
In this section, we describe the following: 1) the problem of registration; 2) the proposed contribution, which uses a multi-scale architecture to improve the U-Net; and 3) the principle of UDGE for unsupervised transfer learning.

\subsection{Problem statement}
Let $X = (x_1, \dots x_{N_X}) \in \mathbb{R}^{N_X \times 3}$ and $Y = (y_1, \dots y_{N_Y}) \in \mathbb{R}^{N_Y \times 3}$ be two point clouds. The goal of registration is to find the right transformation $R \in SO(3), t \in \mathbb{R}^3$, which are the sets of 3D rotations and translations. 
In other words, the goal is to find the set of matches $\mathcal{M}$ and the rotation and translation such that:
{\small
\begin{equation}
    (R^*, t^*, \mathcal{M}^*) = \argmin_{R \in SO(3), t \in \mathbb{R}^3} \sum_{(i, j) \in \mathcal{M}} \|Rx_i +t - y_j\|^2.
\end{equation}
}%

Simultaneously, finding the right match and the right transformation is difficult. This problem is divided into two sub-problems. First, we must find correct matches between point clouds, and second, we compute the transformation using a robust estimator, such as RANSAC~\cite{fischler_bolles_1981}, TEASER~\cite{yang2020teaser}, or FGR~\cite{leibe_fast_2016}. 
%In our work, we will focus on improving the computation of features using U-Net based methods and keep TEASER as robust transformation estimator.
In our work, we focus on using U-Net based methods to improve the computation of the descriptors. 
For the robust transformation estimator, we use TEASER~\cite{yang2020teaser}, because it is faster than RANSAC~\cite{fischler_bolles_1981}. % and as good as RANSAC~\cite{fischler_bolles_1981}.

\paragraph{U-Net-based methods:}
%reference of the unet
%A U-Net architecture is divided into two parts with an encoder, where we down-sample the point cloud and compute feature of larger dimension and a decoder part, where the down-sampled point cloud is up-sampled and fused with features computed on the encoder.
%reference of the unet
%A U-Net architecture is divided into two parts with an encoder, where we down-sample the point cloud and compute feature of larger dimension and a decoder part, where the down-sampled point cloud is up-sampled and fused with features computed on the encoder.
A U-Net architecture can be divided into two parts with an encoder and a decoder; the first part is created by down-sampling the point cloud and computing intermediate features, while the second part is up-sampled and fused with the features computed on the encoder.
Thus, a U-Net instantaneously computes descriptors on all points of a point cloud.
% precise the input ??
Let $\psi_\theta$ be the U-Net neural network of parameter $\theta$. As input, $\psi_\theta$ takes a point cloud (3D coordinates) and input features associated with each point (usually $1$). Therefore, each pair of point clouds $(X \in \mathbb{R}^{N_X \times 3}, Y \in \mathbb{R}^{N_Y \times 3})$ have associated input features ($f^{(in)}_X \in \mathbb{R}^{N_X \times d_{in}}$ and $f^{(in)}_Y \in \mathbb{R}^{N_Y \times d_{in}}$, respectively) with $d_{in}$ the dimension of the input feature. We use $\psi_\theta$ to compute output features $F_{X} \in \mathbb{R}^{N_X \times d}, F_{Y} \in \mathbb{R}^{N_Y \times d}$:
\begin{align}
    F_{X} &= \psi_{\theta}(X, f^{(in)}_X),\\
    F_{Y} &= \psi_{\theta}(Y, f^{(in)}_Y),
\end{align}
where $d$ is the output dimension.

With this formulation, we can express many popular architectures for registration (such as FCGF~\cite{choy2019fully} or D3Feat~\cite{bai2020d3feat}).
Then, finding the right output descriptors is a problem of metric learning. 
We attempt to compute descriptors with minimal distance between positive matches and maximum distance between negatives matches.
For each pair of point clouds $(X, Y)$, we minimize the hard negative contrastive loss:
\begin{align}
    L(\theta) =  &\sum_{(i, j) \in \mathcal{M}^+} \{ [\|F_{X_i} - F_{Y_j}\| - m_+]_{+}^2 \\
    & + \frac{1}{2} [m_{-} - \min_{k|(i, k) \in \mathcal{M}^-} \|F_{X_i} - F_{Y_k}\|]_{+}^2 \\
    & + \frac{1}{2} [m_{-} - \min_{k|(k, j) \in \mathcal{M}^-} \|F_{X_k} - F_{Y_j}\|]_{+}^2 \},
\end{align}
where $[.]_{+} = \max (., 0)$, $\mathcal{M}^+$ is the set of positive matches (ground truth matches), and $\mathcal{M}^-$ is the set of negative matches. $m+$ is a hyper-parameter called the positive margin, and $m-$ is called the negative margin. We kept $m+=0.1$ and $m-=1.4$ as in FCGF~\cite{choy20194d}.\\

Although it is also possible to utilize triplet loss, empirically contrastive loss
 has shown better results~\cite{choy2019fully}. In supervised training, we use the ground truth transformation between pairs of point clouds to obtain positive and negative matches.

\paragraph{Sparse convolution challenges:}
When coupled with the RANSAC estimator, FCGF has shown state-of-the-art results on the 3DMatch dataset. This method can handle large point clouds, is memory efficient, and is faster than most deep methods. But when point clouds come from different sensors or a different environment, FCGF cannot generalize (as noted by Bai \etal~\cite{bai2020d3feat} and Poiesi \etal~\cite{Poiesi2021}). FCGF uses sparse voxel convolutions, so has to voxelize the point cloud with a fixed voxel size. Nevertheless, it seems that sparse convolution can overfit on specific sampling. FCGF, because of this, cannot be applied on a dataset where the sampling is different.
One solution is to downsample the point cloud, %But by downsampling the point cloud, we represent the geometry with fewer points so points with different densities will be approximately similar.
but downsampling leads to a loss of detail and to representations with fewer points. This can impair descriptor matching and lead to points with different densities appearing approximately similar.
%TODO FIGURE figure to show this phenomena
The problem is we also lose details that could improve descriptor matching. Moreover, it is difficult to fix the side length of the voxel.

\subsection{Multi-scale network, MS-SVConv}
% FIGURE Illustration
\begin{figure*}
    \centering
    \includegraphics[width=\textwidth]{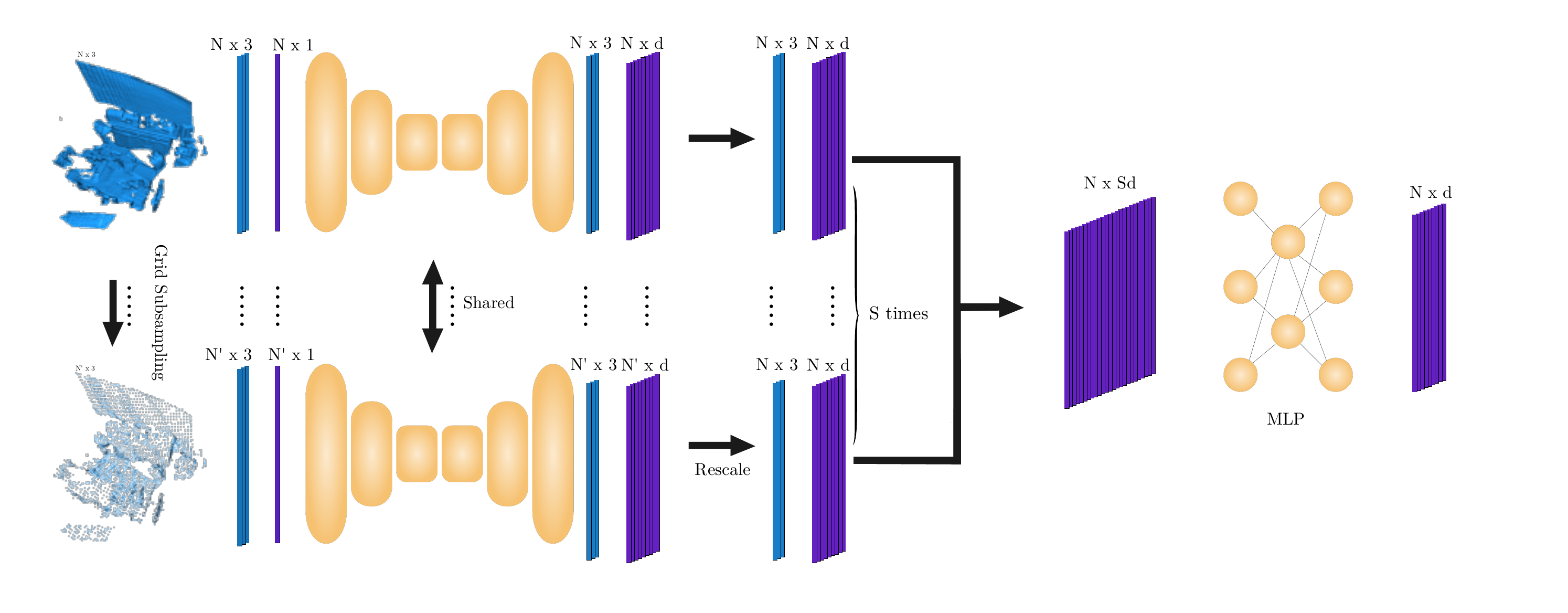}
    \caption{Architecture of the proposed Multi-Scale Sparse Voxel Convolution (MS-SVConv) for registration. The point cloud is downsampled at each scale with a grid subsampling that increases the side length of the voxel by 2 each time.}
    \label{fig:mssvconv}
\end{figure*}
Downsampled point clouds have a homogeneous density but fewer details, whereas highly dense point clouds contain many details, but also marked variations in density across the scene.
%have a lot of details with density varying a lot across the scene.
%Thus, our idea is to create a pyramid of scales with downsampled point clouds.
Thus, we downsample the point cloud at different scales and then apply the U-Net (we will call it a head) on the downsampled point cloud. Finally, we fuse the features computed by each head using a Multi Layer Perceptron (MLP). An illustration of our multi-scale network is shown in Figure~\ref{fig:mssvconv}.
Let $X$ be the input point cloud $X \in \mathbb{R}^{N_X \times 3}$ (and the associated input feature $f^{(in)}_X \in \mathbb{R}^{N_X \times d_{in}}$).
Let $s$ be a scale and $\psi_{\theta}^{(s)}$ a U-Net operating at the scale $s$ with the parameter $\theta$: 
\begin{equation}
\small
    F_X = MLP(\bigoplus_{s=1}^S \psi_{\theta}^{(s)}(X, f^{(in)}_X)),
\end{equation}
where $\bigoplus$ means concatenation. We apply the same MLP for each output of the U-Net.\\
The MLP will learn to select and filter outputs of each scale.
As input, we voxelize the point cloud by increasing the side length of the voxel by two at each scale (performing a grid subsampling). The number of occupied voxels will then be different for each U-Net. 
We assign the same output descriptor for all points of the original point cloud that fall into the same voxel at a specific scale (called Rescale in Figure~\ref{fig:mssvconv}).
%\JED{Feature propagation : tous les points du nuage de points original dans un voxel ont le feature du voxel}

%For each head, we could use a different U-Net with new weights or the same network with shared weights. We will show experiments demonstrating the superiority of a network with weights shared between scales.

In the U-Net architecture, there are multiple downsampling operations that increase the receptive field. However, a classical U-Net for a 3D point cloud with three layers of downsampling brings a receptive field of eight times the side length of the initial voxel. By using three scales of point clouds, we are able to multiply the side length of the initial voxel by 64, bringing more global contexts to computation of the descriptors. 

We use the same network for the different scales with shared weights: this means that we keep the same number of parameters as for one U-Net scale and add only the additional parameters of the final MLP.
%as one scale U-Net with only the additional parameters of the final MLP (less than 10k additional parameters).

%Many multi-scale strategies exist for different tasks. 
Prior works have investigated multi-scale architectures in 3D, but in different ways and contexts. Mu-net~\cite{munet} is an unshared sequential U-Net on dense voxel grid for denoising. This method is different from our parallel shared multi-scale U-Net with sparse convolution.
MS-DeepVoxScene~\cite{roynard2018multiscale} takes several neighborhoods (scales) as input to classify only one point, and is used for semantic segmentation of point clouds.
%taking as input several neighborhoods (scales)
However, in our case, we take the whole point cloud sub-sampled at different scales. The multi-scale network both improves the descriptors in supervised learning, and the capacity of U-Net architectures to generalize.

%MS-SVConv could be used in two ways: for $s$ heads it's possible to have only one Unet that is applied at different scale (called the shared version) but it is also possible to have as many Unet as the number of heads (called the unshared version). The shared version have less parameter and have better results than the unshared version (as shown in the experiment section). So when it is not specified, MS-SVConv refer to the shared version.

\subsection{Unsupervised transfer learning with UDGE}
A supervised setting includes pairs of point clouds and the relative pose between them. The two points clouds are different because they come from different points of view of the scene. The principle of Unsupervised transfer learning with Data GEneration (UDGE) is to use specific data generation to create two partial point clouds from one point cloud. We randomly crop the original data and then apply periodic sampling to simulate partially overlapping views from one scene. This method allows to generate two point clouds, while knowing perfectly the positive and negative matches. 
In the proposed method, we train in a supervised fashion on a source dataset~$S$ and then use UDGE on a target dataset~$T$ by using, as initialization, weights trained on~$S$. We do not need the ground truth poses on target dataset~$T$, and we will show improvements after transfer learning, even if the target dataset~$T$ is small.
%For many registration methods, we need two different views of a scene or an object. It's possible because either we have the complete object (like in ModelNet) or because we have different views of a scene with their respective ground truth pose. Our goal is to provide an unsupervised way of training even if we only have partial scans without ground truth pose. 
The work closest to UDGE is from~\cite{yuan2020selfsupervised}, but the authors performed self-supervision without pre-training. We show that without pre-training, self-supervision can work for large datasets but not at all for small datasets. 
%Then they use the same point cloud as a pair while we have a specific procedure like the crop and the periodic sampling to create two partial overlapping point clouds.
Additionally, the author used only one point cloud as a pair, while we utilize specific procedures, like cropping and periodic sampling, to create two partially overlapping point clouds.
% FIGURE and Algorithm

\begin{figure}
    \centering
    \includegraphics[width=\linewidth]{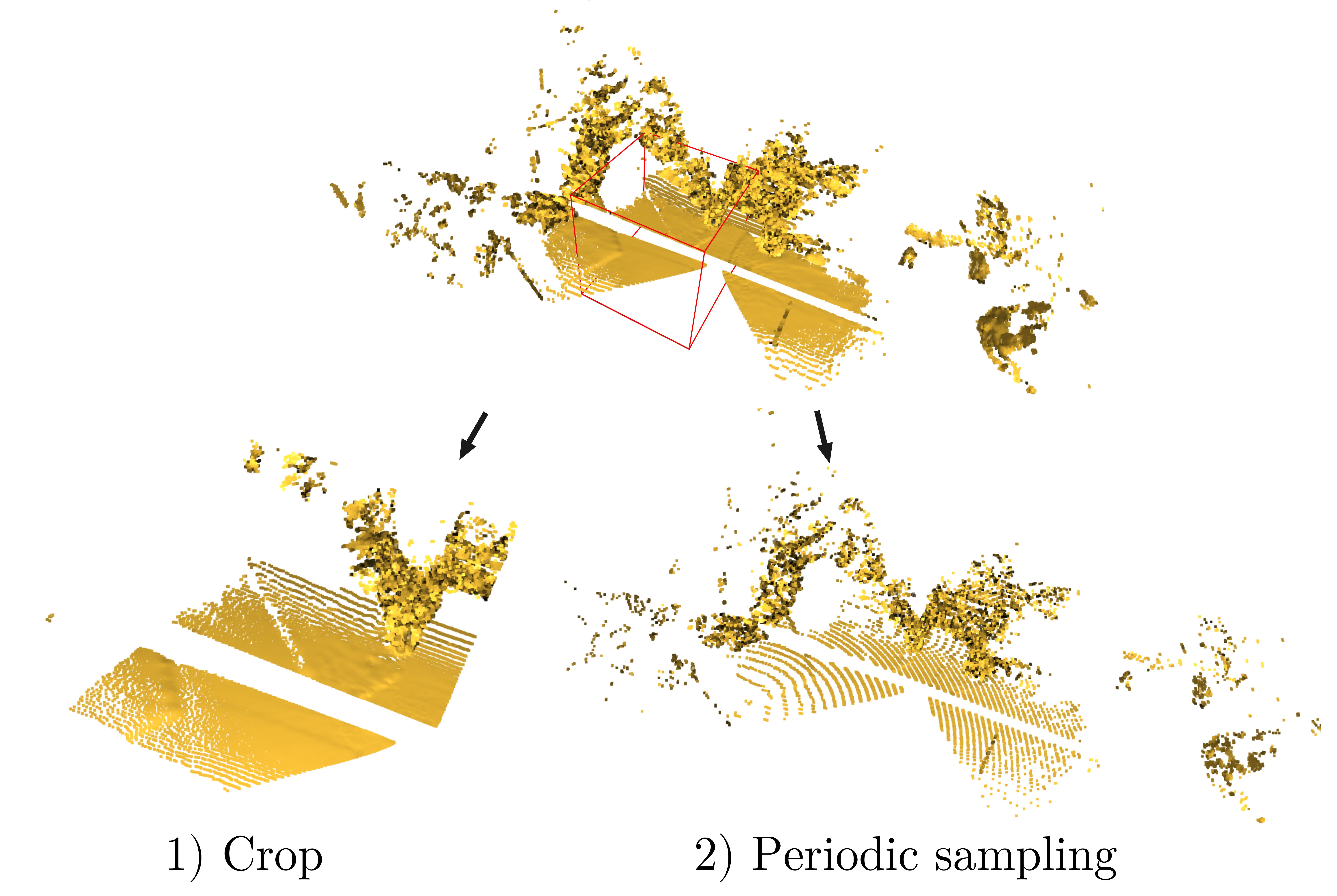}
    \caption{The two data generation methods we use in our unsupervised transfer learning.}
    \label{fig:pipeline}
\end{figure}

\paragraph{Our data generation:}
%In a new dataset, we do not have pairs of point cloud and we will create them from each point cloud.
%But it's not always possible to have this relative pose for different reason. We propose an unsupervised method to generate two different point clouds from one point cloud. We call it data generation.
%We propose two data generation procedure to generate different point cloud from one point cloud.
Figure~\ref{fig:pipeline} shows how data is generated from one original scan. We distinguish between data generation and data augmentation. Data generation concerns the proposed process of making two partial scans from a point cloud. Thus, the goal of data generation is to generate, without supervision, new data from a point cloud that is in the same case as data from supervised learning. 
%data generation's goal is to generate without supervision new data from one point cloud to be in the same case as supervised learning. 
The data generation parameters depend on the target dataset. In contrast, data augmentation is used to artificially increase the size of the dataset. In our case, we also perform data augmentation with the two scans, in particular the classical data augmentation in point clouds that are random rotation around all axes, random scale and noise (jitter).
Data generation is already used on ModelNet to simulate partial views of an object, but it is mainly used to test methods on partial scans (see~\cite{yew2020rpmnet}, for example). It is not used for training. In our case, we apply our strategy on real unknown datasets from RGB-D frames or LiDAR scans. 
Although a similar work is undertaken in~\cite{zhang_depth_contrast}, they performed
%We can also see close work in~\cite{zhang_depth_contrast} but they do
self-supervision only for pre-training for semantic segmentation.
In the case of UDGE, we use data generation for unsupervised transfer learning for point cloud registration and not as pre-training.

\paragraph{Crop:}
Similar to images, it is possible to select a local zone of the point cloud and learn from it.
To accomplish this, we select a random point and a random 3D shape (it can be a cube or a sphere). Points inside the shape are kept, and points outside the shape are discarded. Selecting a local zone is very useful for simulating pairs of partially overlapping point clouds.

%Figure to show data augmentation.

\paragraph{Periodic sampling:}
Real scans do not necessarily have regular sampling because sampling depends on the point of view. We propose a data generation technique to simulate irregular sampling that is called periodic sampling (see in Figure~\ref{fig:pipeline}). The principle is to remove points periodically with respect to a center chosen randomly.
Let $c \in \mathbb{R}^3$  be a random center. We can compute the mask $M \in \{0, 1\}^N$of the point cloud $X \in \mathbb{R}^{N \times 3}$. $m_i$ is a binary value that indicates whether we keep $x_i$ or not. The proposed mask is:
\begin{equation}
\small
    m_i = \mathbf{1}(|\cos (\frac{2 \pi}{T}\|x_i-c\|) | > \cos (\alpha \pi)),
\end{equation}
where $\alpha \in [0, 1]$ is a threshold that indicates the proportion of points we want to keep. If $\alpha=0$, every point is removed, and if $\alpha=1$, we keep every point. $T$ is the periodic sampling period. By changing the period $T$ and the threshold $\alpha$, we can simulate a wide range of different types of sampling.
Periodic sampling is especially useful for scenes, where sampling is highly irregular.
% TODO image of periodic sampling

\section{Experiments}

We evaluate our contributions on the 3DMatch~\cite{zeng20163dmatch}, ETH~\cite{Pomerleau:2012} and TUM~\cite{sturm12iros} datasets. 
We first describe the datasets used for the experiments, and we show the impact of the multi-scale architecture on supervised learning. Finally, we show the results for generalization of MS-SVConv thanks to multi-scale architecture and UDGE. 
The implementation details and more experiments on MS-SVConv and UDGE can be found in the supplementary material.

\subsection{Datasets}
\paragraph{ModelNet~\cite{Wu_2015_CVPR} (source dataset):}

ModelNet40~\cite{Wu_2015_CVPR} is a CAD dataset containing around 10,000 objects with 40 object categories. This dataset is used for classification, but many deep learning methods use it for point cloud registration~\cite{Wang_2019_NeurIPS, aoki2019pointnetlk, wang2019deep}. We use the training set of ModelNet40 for pre-training. 
Our goal is to show that we can train on a synthetic dataset, such as ModelNet~\cite{Wu_2015_CVPR}, and can generalize on a real-world dataset, such as the ETH dataset.

\paragraph{3DMatch~\cite{zeng20163dmatch} (source and target dataset):}
3DMatch~\cite{zeng20163dmatch} is a dataset composed of RGB-D frames from different indoor datasets related to 3D reconstruction,
%dataset of RGB-D frames from different indoor datasets of 3D reconstruction,
such as BundleFusion~\cite{dai2017bundlefusion}, SUN3D~\cite{HalberF2C17sun3D}, or 7-Scene~\cite{shotton20137scene}. The dataset consists of 62 scenes, and as in~\cite{choy2019fully}, we split the dataset into training with 48 scenes, validation with 6 scenes, and a test set of 8 scenes. 
%Thus, to create point clouds, we need to process the depth frames. To generate them, we use TSDF fusion from 50 depth images (as in~\cite{deng2018ppfnet, choy2019fully, bai2020d3feat, gojcic2018perfect}).
To create point clouds, we must first process the depth frames; to generate these point clouds, we use TSDF fusion from 50 depth images~\cite{deng2018ppfnet, choy2019fully, bai2020d3feat, gojcic2018perfect}.
The test set is provided by the 3DMatch website. 3DMatch~\cite{zeng20163dmatch} is used for supervised training (pairs of fragments with 30\% overlap as in~\cite{gojcic2018perfect}). We also use 3DMatch as the target dataset for UDGE with pre-training on ModelNet.

\paragraph{ETH Dataset~\cite{Pomerleau:2012} (target dataset):}
The ETH dataset~\cite{Pomerleau:2012} is an outdoor and indoor 3D point cloud dataset acquired with a 2D LiDAR: it is composed of eight scenes, six outdoors and two indoors. 
%ETH is too small to be split contrary to 3DMatch~\cite{zeng20163dmatch},
%Unlike 3DMatch~\cite{zeng20163dmatch}, ETH is too small to be split.
%So it is generally used to test generability. 
This dataset is considered 
to be difficult to work with because of complications related to noise and irregular density.
%as a difficult because of its scans with noise and irregular density. 
%We apply unsupervised transfer learning on our network for ETH dataset~\cite{Pomerleau:2012}, compare with other methods, and use the dataset prepared by Fontana~\etal~\cite{fontana2020benchmark} to evaluate our results. 
%For ETH, the evaluation protocol that we have chosen is different from previous papers. 
We will make a difference between the ETH 8-scenes, following the rigorous protocol described by Fontana \etal~\cite{fontana2020benchmark} and the ETH 4-scenes, following the Gojcic \etal's benchmark~\cite{gojcic2018perfect} and mainly followed in previous published works such as~\cite{bai2020d3feat, ao2020SpinNet, Poiesi2021, gojcic2018perfect, Poiesi2021gedi}.

\paragraph{TUM dataset \cite{sturm12iros} (target dataset):}
TUM is a RGB-D dataset~\cite{sturm12iros} mainly used for RGB-D SLAM or odometry. We use a single frame as point cloud, following the protocol of Fontana \etal~\cite{fontana2020benchmark}. This dataset is also used to evaluate the proposed unsupervised transfer learning method.

\subsection{Metrics}
To evaluate the method, we use two metrics: the Feature Match Recall (FMR) and the Scaled Registration Error (SRE).

Let $(X_i, Y_i)$ be a pair of scenes. 
We call $\mathcal{M}(X_i, Y_i)$ the matches between $X_i$ and $Y_i$, $|\mathcal{M}|$ the number of matches, and $(R^{(gt)}, t^{(gt)})$ the ground truth rotation and translation.
The hit ratio is:
\begin{equation}
\small
    H(X_i, Y_i) = \frac{1}{|\mathcal{M}|} \sum_{k, l \in \mathcal{M}(X_i, Y_i)}\mathbf{1}(\|R^{(gt)} x_k + t^{(gt)} - y_l\| \leq \tau_1).
\end{equation}

The Feature Match Recall (FMR) is then defined as
\begin{equation}
\small
    \text{FMR} = \frac{1}{N}\sum_{i=1}^N \mathbf{1}(H(X_i, Y_i)  \geq \tau_2),
    \label{equ:feat_match_recall}
\end{equation}
where $N$ is the number of pairs of scenes. As in previous works~\cite{gojcic2018perfect, bai2020d3feat, choy2019fully, Poiesi2021}, we assign $\tau_1=0.1$~m and $\tau_2=0.05$ by default. Similar to~\cite{gojcic2018perfect, bai2020d3feat, ao2020SpinNet, Poiesi2021, Li_2020_CVPR}, we perform a symmetric test to filter the matches before evaluation. This metric allows evaluating descriptor matching. Although it is useful for comparison of different descriptor-matching methods, we cannot use it to compare different registration algorithms (and therefore, we cannot use it to evaluate classical registration algorithms, such as ICP).
To measure the registration error, we use the error defined by Fontana \etal~\cite{fontana2020benchmark}. We call it the Scaled Registration Error (SRE).
Suppose we have a point cloud $X \in \mathbb{R}^{N_X \times 3}$ and $(R^{(gt)}, t^{(gt)})$ is the ground truth transformation between $X$ and $Y$. We want to evaluate our algorithm, which produces the transformation $(R^*, t^*)$.
The SRE for a pair X, Y is defined as:
{\scriptsize
\begin{align}
    SRE(X, Y) &= \frac{1}{N_X} \sum_{i=1}^{N_X} \frac{\|(R^{(gt)} x_i + t^{(gt)}) - (R^* x_i + t^*)\|}{\|(R^{(gt)} x_i + t^{(gt)}) -  (R^{(gt)} \bar{x} + t^{(gt)}) \|} \\
    \bar{x} &= \frac{1}{N_X} \sum_{i=1}^{N_X} x_i.
\end{align}
}%
The SRE depends on X and Y, because we estimate $R^*$ and $t^*$ from $X$ and $Y$.
For every pair of scans in a dataset with $N$ pairs, we compute the median of the SRE:
\begin{equation}
    SRE = \text{median}_{i=1\dots N}(SRE(X_i, Y_i)).
\end{equation}
The mean is sensitive to outlier results and is not representative of the results, as explained in~\cite{fontana2020benchmark}.

\subsection{Evaluation of supervised learning}
\begin{table}[ht]
\small
\centering
\begin{tabular}[t]{lcc}
\toprule
\multicolumn{3}{c}{\textbf{Supervised learning on 3DMatch}} \\
\midrule
 & FMR (\%) & FMR (\%) \\
Methods & $\tau_2=0.05$ & $\tau_2=0.2$\\
\midrule

SHOT~\cite{Salti2014SHOTUS} & 23.8 & - \\
FPFH~\cite{rusu_fast_2009} & 35.9 & - \\
3DMatch~\cite{zeng20163dmatch} & 59.6 & - \\
PPFNet~\cite{deng2018ppfnet} & 62.3 & - \\
3DSmoothNet~\cite{gojcic2018perfect} & 94.7 & 72.9\\
DIP~\cite{Poiesi2021} & 94.8 & - \\
FCGF~\cite{choy2019fully} & 95.2 & 67.4 \\
FCGF$^*$~\cite{choy2019fully} & 97.5 & 87.3 \\
D3Feat~\cite{bai2020d3feat} & 95.8 & 75.8 \\
Multiview~\cite{Li_2020_CVPR} & 97.5 & 86.9 \\
SpinNet~\cite{ao2020SpinNet} & 97.6 & 85.7 \\
Predator~\cite{huang2020predator} & 96.6 & - \\
GeDI~\cite{Poiesi2021gedi} & 97.9 & - \\
MS-SVConv (1 head) & 97.6 & 87.2 \\
MS-SVConv (3 heads) & \bf{98.4} & \bf{89.9}\\
\bottomrule

\end{tabular}
\caption{Feature Match Recall (FMR) on 3DMatch for two hit ratio parameters $\tau_2$. Results from published methods are taken from the papers. FCGF$^*$ means that we evaluate ourselves the original code with a symmetric test, before computing the FMR.}
\label{tab:resultsSupervised}
\end{table}%
We evaluate the impact of MS-SVConv on supervised learning to see the influence of the multi-scale architecture.
Table~\ref{tab:resultsSupervised} shows the results on 3DMatch. Although the 3DMatch benchmark is very competitive, MS-SVConv outperforms all published methods. If we compare MS-SVConv with the best published method (MultiView~\cite{Li_2020_CVPR}), we have an augmentation of +3\% on the Feature Match Recall (FMR) with $\tau_2 = 0.2$. Table~\ref{tab:resultsSupervised} shows that MS-SVConv with three heads is much better than with a single head (a single scale), which demonstrates the interest of multi-scale methods for supervised learning. MS-SVConv with one head and FCGF are conceptually similar but MS-SVConv has a different implementation, see supplementary material. Moreover, contrary to other methods, FCGF does not filter matches with a symmetric test. If we filter matches with a symmetric test, results of FCGF and MS-SVConv with one head are similar (see Table ~\ref{tab:resultsSupervised}).

\subsection{Evaluation of UDGE}

%\begin{table}[ht]
%\centering
%\tabcolsep=0.11cm
%\begin{tabular}[t]{lcccc}
%\toprule
%\multicolumn{5}{c}{ETH Dataset} \\
%\midrule
%Methods & Voxel (cm) & FMR (\%) & SRE & Time (s)\\
%\midrule
%\multicolumn{5}{c}{Classical methods} \\
%Point2Point ICP & - & - &2080 & 0.25\\
%Point2Plane ICP & - & - &2110 & 0.38\\
%FPFH~\cite{rusu_fast_2009} & - & 0.75 & 55 & 0.71 \\
%\midrule
%\multicolumn{5}{c}{Deep methods without UDGE} \\
%&0.983&0.020&0.044 \\
%Multiview~\cite{Li_2020_CVPR} & - & 42.6 & 12 & 56 \\
%D3Feat~\cite{bai2020d3feat} & 6 & 64.5 & 12 & 0.43 \\
%DIP~\cite{Poiesi2021} & - & \bf{93.9} & 6.9 & 8.27   \\
%MS-SVConv(1) & 2 & 56.4 & 13 & 0.24  \\
%MS-SVConv(3) & 2, 4, 8 & 76.8 & 9.6 & 0.52 \\

%\midrule
%\multicolumn{5}{c}{With UDGE} \\
%MS-SVConv(1) & 4 & 87.5 & 7.2 & \bf{0.16} \\
%MS-SVConv(3) & 4, 8, 16 & 93.6 & \bf{6.6} & 0.40 \\
%\bottomrule
%\end{tabular}
% S
%\caption{Feature Match Recall (FMR) and Scaled Registration Error (SRE) x1000 on ETH dataset with deep methods trained on 3DMatch. MS-SVConv(1) means MS-SVConv with 1 head; MS-SVConv(3) means MS-SVConv with 3 heads.  We only report the average time of feature extraction. Results from all methods are computed using codes available online (we kept the best voxel size for each method).}
%\label{tab:unsupervised}
%\end{table}%

\begin{table}[ht]
\centering
\small
\tabcolsep=0.11cm
\begin{tabular}[t]{lc|ccc}
\toprule
\multicolumn{5}{c}{\textbf{Unsupervised learning on ETH dataset}} \\
\midrule
& \textbf{ETH} & \multicolumn{3}{c}{\textbf{ETH}} \\
& \textbf{4-scenes} & \multicolumn{3}{c}{\textbf{8-scenes}} \\
\midrule
Methods & FMR (\%) & FMR (\%) & SRE & Time (s)\\
\midrule
\multicolumn{5}{c}{Classical method} \\
%Point2Point ICP & - & - &2080 & 0.25\\
%Point2Plane ICP & - & - &2110 & 0.38\\
FPFH~\cite{rusu_fast_2009} & 22.1 & 0.75 & 85.1 & 0.71 \\
%SHOT~\cite{Salti2014SHOTUS} & 61.1 & - & - & - \\
\midrule
\multicolumn{5}{c}{Patch based deep methods without UDGE} \\
%&0.983&0.020&0.044 \\
3DSmoothNet~\cite{gojcic2018perfect} & 79.0 & - & - & - \\
Multiview~\cite{Li_2020_CVPR} & 92.3 & 42.6 & 44.0 & 56.0 \\
DIP~\cite{Poiesi2021} & 92.8 & \bf{93.9} & \textbf{6.9} & 8.27   \\
GeDI~\cite{Poiesi2021gedi} & 98.2 & - & - & - \\
\midrule
\multicolumn{5}{c}{U-Net based deep methods without UDGE} \\
D3Feat~\cite{bai2020d3feat} & 61.6 & 64.5 & 95.0 & 0.43 \\
MS-SVConv(1) & 34.9 & 56.4 & 151.0 & 0.24  \\
MS-SVConv(3) & 71.8 & 76.8 & 82.2 & 0.52 \\

\midrule
\multicolumn{5}{c}{U-Net based deep methods with UDGE} \\
MS-SVConv(1) & 88.0 & 87.5 & 44.0 & \bf{0.16} \\
MS-SVConv(3) & \textbf{98.9} & 93.6 & \bf{6.9} & 0.40 \\
\bottomrule
\end{tabular}
% S
\caption{Feature Match Recall (FMR) and median Scaled Registration Error (SRE) x1000 on the ETH dataset with deep methods trained on 3DMatch. ETH 4-scenes follows Gojcic \etal's benchmark~\cite{gojcic2018perfect}, and results from other methods are published results (only FMR available). ETH 8-scenes follows Fontana \etal~\cite{fontana2020benchmark} protocol, and all methods are computed using codes available online (FMR, SRE, and Time). MS-SVConv(1) means MS-SVConv with one head; MS-SVConv(3) means MS-SVConv with three heads.  We report only the average time of descriptor extraction. }
\label{tab:unsupervised}
\end{table}%

Table~\ref{tab:unsupervised} shows the results on the ETH dataset~\cite{Pomerleau:2012} with and without UDGE after pre-training on 3DMatch. ETH is a challenging dataset because of the density variation and missing areas. Without UDGE, MS-SVConv with one head has average results on ETH of only 34.9\% on the ETH 4-scenes and 56.4\% FMR on ETH 8-scenes. MS-SVConv with three heads has +36.9\% FMR on ETH 4-scenes and +20.4\% FMR improvement on ETH 8-scenes without UDGE, thanks to the multi-scale architecture. This demonstrates the capability of multi-scale architecture to improve the generalization capacity of U-Net, and proves useful for registration in online settings.

With UDGE, MS-SVConv(3) gets state-of-the-art results with 98.9\% FMR on ETH 4-scenes and 93.6\% FMR on ETH 8-scenes like the best patch-based method DIP~\cite{Poiesi2021} while 20 times faster. There is a synergy between multi-scale and UDGE, with +10.9\% and +6.1\% improvement of the FMR between one and three heads. %unsupervised transfer learning can be used for offline registration such as 3D reconstruction.

Adding heads increase the generalization capability, but the training time and the inference time are multiplied by 2.5 (from 0.16~s to 0.40~s for the average time of descriptor extraction). Nonetheless, the proposed network is still much faster than any other deep method we tested.

\begin{table*}[ht]
\small
\centering
\begin{tabular}[t]{l|cccc|cccc|c}
\toprule
\multicolumn{10}{c}{\textbf{Unsupervised learning on ETH dataset}} \\
\midrule
 & \multicolumn{4}{c|}{Scenes used for UDGE} & \multicolumn{4}{c|}{Scenes for testing} & \\
Methods & Haupt. & Stairs & Plain & Apart. & Gaz. Sum. & Gaz. Wint.  & Wood Aut. & Wood Sum. & Average \\
\midrule
MS-SVConv(1) & 53 & 93 & 76 & 100 & 85 & 99 & 84 & 89 & 84.9 \\
MS-SVConv(3) & 72 & 99 & 88 & 100 & 96 & 100 & 99 & 99 & 94.1 \\
\bottomrule
\end{tabular}
\caption{Feature Match Recall (FMR) with $\tau_2=0.05$ per scene on the ETH 8-scenes dataset~\cite{fontana2020benchmark}. MS-SVConv was pre-trained on ModelNet, and data generation for UDGE  comes only from Hauptgebaude, Stairs, Plain and Apartment.} 
\label{tab:ethsplitfontana}
\end{table*}%

\paragraph{Influence of the source dataset $S$:}
\begin{table}[ht]
\centering
\small
\begin{tabular}[t]{lcc}
\toprule
\multicolumn{3}{c}{\textbf{Unsupervised learning on ETH dataset}} \\
\midrule
 & FMR (\%) & FMR (\%) \\
\textbf{Source} & (without UDGE) & (with UDGE)\\
\midrule
ModelNet & 74.1 & 93.4  \\
3DMatch & 76.8 &  93.6 \\
\bottomrule
\end{tabular}
\caption{Influence of source dataset $S$ for the target ETH 8-scenes dataset with MS-SVConv(3).}
\label{tab:pretrained}
\end{table}%

Surprisingly, thanks to the proposed multi-scale architecture and the data generation for transfer learning, ModelNet pre-training (48~h instead of three weeks for training on 3DMatch) is enough to obtain very good results with the ETH dataset (see Table~\ref{tab:pretrained}). 
%This shows that the source dataset does not have so much influence on the results of the target dataset.
%This shows that the source dataset and the target dataset can be very different and MS-SVConv will still work
This shows that even if the source dataset and the target dataset are very different, MS-SVConv will still work.
\begin{table}[ht]
\centering
\small
\begin{tabular}[t]{llccc}
\toprule
& \textbf{Target} & ETH & TUM & 3DMatch \\
\textbf{Source} & &   &  &  \\
\midrule
$\emptyset$ & & 0.0 & 16.0 & 60.3 \\
ModelNet &  & 93.4 & 100 & 96.5 \\
3DMatch &  & 93.6 & 99.7 & 97.8 \\
\bottomrule
\end{tabular}
\caption{UDGE on the ETH, TUM, and 3DMatch datasets with no pre-training ($\emptyset$) or with pre-training on ModelNet or pre-training on 3DMatch. The results are Feature Match Recall in \% with MS-SVConv(3).}
\label{tab:transfer}
\end{table}%

Pre-training in a supervised fashion is essential. We can see in Table~\ref{tab:transfer} that, with no pre-training, UDGE's results are very poor, especially on small datasets like ETH and TUM. Decent results (60.3\% FMR) are still produced when the dataset is much larger like 3DMatch. However, supervised pre-training on ModelNet is enough to have state-of-the-art results thanks to UDGE and without any ground truth for any of the target datasets (ETH, TUM, or 3DMatch). It shows that UDGE can bring good results even if the target dataset is small.

\paragraph{Does UDGE allow for generalization on unseen scenes?}

In all experiments, UDGE involves taking the test set and artificially creating point cloud pairs to fit the network to the new dataset. 
Even if we use the test set for the transfer, the proposed protocol is valid because UDGE does not use the ground truth poses of the test set. 
%It can perfectly be used on an unknown dataset. 
But to what extent will the transfered descriptors generalize on unseen scenes? 
To see if the results remain similar, we split the ETH dataset in two: After pre-training on ModelNet, we apply UDGE on MS-SVConv with four scenes from ETH dataset (Plain, Stairs, Hauptgebaude and Apartment) and we evaluate it on the four others (Gazebo Summer, Gazebo Winter, Wood Autumn, and Wood Summer). We use the same hyper-parameters as in the previous training (see the section Implementation Details in supplementary material), except that the number of epochs is 400 instead of 200. Table~\ref{tab:ethsplitfontana} shows that even if MS-SVConv has never seen Gazebo and Wood, the method can still generalize on these scenes. 
%It demonstrates that UDGE can transfer the property of a dataset without 

\paragraph{Influence of the proposed data generation:}
\begin{table}[ht]
\centering
\small
\begin{tabular}[t]{ccc}
\toprule
Crop & Periodic Sampling & FMR (\%) \\ % & SRE \\
\midrule
& & 88.4 \\ %& 7.6\\
\checkmark& & 91.4 \\ %& 6.7\\
\checkmark&\checkmark&\bf{93.4} \\ %& \bf{6.6}\\
\bottomrule
\end{tabular}
\caption{Influence of the proposed data generation in UDGE: MS-SVConv(3) is pre-trained on ModelNet and UDGE is applied on the ETH 8-scenes dataset.}
\label{tab:augment}
\end{table}%
We performed experiments on the ETH dataset to see the influence of the proposed data generation of UDGE. After pre-training MS-SVConv(3) on ModelNet, we tried UDGE on the ETH dataset without cropping or using periodic sampling (using the same point cloud for pairs as in~\cite{yuan2020selfsupervised}). Table~\ref{tab:augment} shows that cropping is important in data generation. If we do not perform this operation, the performance drops from 91.4\% to 88.4\% on the FMR. Periodic sampling also brings improvements. This experiment highlights that the data generation method in UDGE improves the registration results.

\section{Conclusion}
We presented MS-SVConv, a multi-scale U-Net based method for descriptor matching in point cloud registration. We also presented UDGE, a simple yet efficient method for generalizing on unknown scenes.
%We presented MS-SVConv a multi-scale U-Net based method for feature matching for the task of point cloud registration.
The multi-scale architecture leads to gain state-of-the-art results on 3DMatch for supervised learning. With UDGE, it is possible to transfer descriptors for registration on an unknown dataset without any supervision. Additionally, simply by pre-training on a synthetic dataset like ModelNet, we obtain state-of-the-art results on the ETH, TUM, and 3DMatch datasets with UDGE. In contrast to patch-based methods, 
%our method keep the advantages of U-Net based method which are speed, efficiency and modularity. 
MS-SVConv retains the simplicity, speed, efficiency, and modularity of U-Net-based methods.
%Moreover, our method could be applied in many contexts or it could be applied for different tasks such as semantic segmentation.
%Moreover, we believe our multi-scale architecture could be used as efficient backbone in other tasks like semantic segmentation and object detection for point clouds.
%Moreover, we believe that our multi-scale architecture can, in the future, function as the backbone of other tasks, including semantic segmentation and object detection for point clouds.
\paragraph{Acknowledgments}
This work was granted under the funding of the Idex PSL with the reference ANR-10-IDEX-0001-02 PSL. This work was granted access to the HPC resources of IDRIS under the allocation 2020-AD011012181 made by GENCI.

{\small
\bibliographystyle{ieee_fullname}
\bibliography{egbib}
}

%%%%%%%%%%%%%%%%%%%%%%%%%%%%%%%%%%%%%%%%%%%%%%%%%%%%%%%%%%%%%%%%% SUP %%%%%%%%%%%%%%%%%%%%%%%%%%%%%%%%%%%%%%%%%%%%%%%%%%%%%%ù
\appendix
\title{Supplementary material: 3D Point Cloud Registration with Multi-Scale Architecture and Unsupervised Transfer Learning}
\author{Sofiane Horache
\and Jean-Emmanuel Deschaud 
\and François Goulette\\
MINES ParisTech, PSL University, Centre for Robotics, 75006 Paris, France\\
\small{firstname.surname@mines-paristech.fr}\\
}
\maketitle
We present additional analysis of the method presented in the main article with justifications of the design choices (hyper-parameters of the U-Net network, influence of the number of heads, importance of shared weights) as well as a more in-depth analysis of the results on the 3DMatch and ETH datasets. 
The first section describes the experiments on MS-SVConv, and the multi-scale architecture in the supervised case. The second section focuses on experiments on UDGE, and the synergy between UDGE and MS-SVConv.
The third section specifies implementation and protocol details. 
Finally, we show qualitative results on the last section (with images) of the registration on the 3DMatch, the ETH, and the TUM datasets.\\

%ETH : Hauptgebaude, Plain, Stairs, Apartment, Gazebo Summer, Gazebo Winter, Wood Summer and Wood Autumn

\section{More experiments on supervised learning with MS-SVConv}
In this section, we present more experiments to show that MS-SVConv with three heads has great generalization capabilities in comparison with MS-SVConv with one head. We show that a simple pre-training on ModelNet can bring significant results. We also present an experiment to show that MS-SVConv is robust to random rotations. We also show that even if the number of interest points for the matching is low, MS-SVConv have great results on the 3DMatch dataset. Finally, we show that MS-SVConv works very well on scenes with low overlap. The last subsection shows details of results of the 3DMatch dataset

\subsection{Generalization capability of MS-SVConv without UDGE}

\begin{table}[ht]
\small
\centering
\begin{tabular}[t]{llccc}
\toprule
& \textbf{Target} & ETH & TUM & 3DMatch \\
\textbf{Architecture} & &  &  &  \\
\midrule
MS-SVConv(1) & & 56.4 &99.0 & 97.6 \\
MS-SVConv(3) & & \bf{76.8} & \bf{100} & \bf{98.4} \\
\bottomrule
\end{tabular}
\caption{Feature Match Recall ($\tau_2=0.05$) of MS-SVConv trained on the 3DMatch dataset with a voxel size of 2 cm and evaluated on ETH 8-scenes, TUM and 3DMatch without UDGE.}
\label{tab:transfer4}
\end{table}%

\begin{table}[ht]
\small
\centering
\begin{tabular}[t]{llccc}
\toprule
& \textbf{Target} & ETH & TUM & 3DMatch \\
\textbf{Architecture} & &  &  &  \\
\midrule
MS-SVConv(1) & & 37.4 &28.3 & 29.2 \\
MS-SVConv(3) & & \bf{74.1} & \bf{99.3} & \bf{85.0} \\
\bottomrule
\end{tabular}
\caption{Feature Match Recall ($\tau_2=0.05$) of MS-SVConv trained on the ModelNet dataset with a voxel size of 2 cm and evaluated on ETH 8-scenes, TUM and 3DMatch without UDGE.}
\label{tab:transfer3}
\end{table}%

Table~\ref{tab:transfer4} shows the Feature Match Recall without UDGE on ETH 8-senes, TUM and 3DMatch datasets after having trained our models on the 3DMatch dataset (Table~\ref{tab:transfer3} when we train on the ModelNet dataset). These tables show that multi-scale improves generalization capabilities. Moreover, when the training set is the ModelNet dataset, Table~\ref{tab:transfer3} shows that multi-scale brings huge improvement (up to +71\%) between MS-SVConv with three heads and our U-Net with one head. With the proposed multi-scale sparse voxel convolutions, ModelNet as a pre-training set is enough to compute meaningful descriptors.

\subsection{Robustness to random rotations}

\begin{table}[ht]
\small
\centering
\begin{tabular}[t]{lrcc}
\toprule
&  & w/o Rotation & w Rotation  \\
Architecture & &  &  \\
\midrule
MS-SVConv(1) & & 87.2 & 87.0  \\
MS-SVConv(3) & & 89.9 & 89.7  \\
\bottomrule
\end{tabular}
\caption{Feature Match Recall on 3DMatch ($\tau_2=0.2$) in supervised learning with and without random rotations in the test set.}
\label{tab:rotated}
\end{table}%

U-Net based methods with sparse voxel convolutions are often considered non-robust to random rotations. It is true that they are not invariant by rotation by design, but become robust thanks to data augmentation. In all of our training, we added data augmentation with random rotations in all directions. To test the final robustness, we added large random rotations in the 3DMatch test set. Table~\ref{tab:rotated} shows that the results are almost equivalent on the 3DMatch dataset with or without rotation.

\subsection{Influence of the number of points}
For all the previous experiments on the 3DMatch and the ETH dataset, we sample randomly 5000 points before matching. To show that MS-SVConv computes meaningful descriptors for each points, we tried to sample different number of points. Table~\ref{tab:num_pt_3dm} shows that even if we sample 250 points, we have a FMR of 96.3 so a loss of 2.1\%. Except the patch-based mathod GeDI, all other methods have a more significant drop of result when the number of point decreases. It shows that MS-SVConv computes more meaningful descriptors at each points, thanks to the multi-scale computation that brings more contexts.
\begin{table}[ht]
\scriptsize
\centering
\begin{tabular}[t]{lcccccc}
\toprule
Number of points & 5000 & 2500 & 1000 & 500 & 250 & Av.\\
\midrule
Perfect Match~\cite{gojcic2018perfect} & 94.7 &94.2 &92.6 &90.1 &82.9 &90.9 \\
FCGF~\cite{choy2019fully} & 95.2& 95.5 & 94.6 & 93.0 & 89.9 & 93.6 \\
D3Feat(rand)~\cite{bai2020d3feat} & 95.3 &95.1 &94.2& 93.6& 90.8& 93.8 \\
D3Feat(pred)~\cite{bai2020d3feat} & 95.8 & 95.6& 94.6& 94.3& 93.3& 94.7 \\
SpinNet~\cite{ao2020SpinNet} & 97.6 &97.5& 97.3& 96.3& 94.3& 96.6 \\
GeDI~\cite{Poiesi2021gedi} & 97.9 & 97.7& \bf{97.6}& 97.2& \bf{97.3}& \bf{97.5} \\
\midrule
MS-SVConv(3) & \bf{98.4} & \bf{97.9} & 97.0 & \bf{97.9} & 96.3 & \bf{97.5} \\
\bottomrule
\end{tabular}
\caption{Influence of the number of points for supervised learning on the 3DMatch dataset (FMR with $\tau_2=0.05$).}
\label{tab:num_pt_3dm}
\end{table}%

\subsection{Results on the 3DLoMatch dataset}
3DLoMatch~\cite{huang2020predator} is a dataset containing the previously ignored scan pairs of 3DMatch that have low 10-30\% overlap.
We compare MS-SVConv with Predator~\cite{huang2020predator} on the 3DMatch and 3DLoMatch dataset. Predator is a very sophisticated method that use graph convolutional neural network and deep attention module after the encoder to compute meaningful descriptors even when the overlap is very low. On the contrary, MS-SVConv uses a simple multi-scale architecture.
We compare MS-SVConv(3) on 3DMatch and 3DLoMatch. Table~\ref{tab:fmr_lo_3dm} shows that MS-SVConv outperforms Predator on 3DMatch and 3DLoMatch (please, note that contrary to Predator, MS-SVConv does not use a detector to select keypoints, but take random points).
%From this experiment, we can conclude that a simple multi-scale solution is more efficient to deal with scenes with a low overlap than an elaborate architecture. 
\begin{table}[ht]
\small
\centering
\begin{tabular}[t]{lcc}
\toprule
 & 3DMatch & 3DLoMatch\\
\midrule
Predator~\cite{huang2020predator} & 96.6 & 73.0 \\
MS-SVConv(3) & \bf{98.4} & \bf{77.2}\\
\bottomrule
\end{tabular}
\caption{Comparison between Predator~\cite{huang2020predator} and MS-SVConv(3) on the 3DMatch and 3DLoMatch dataset in supervised learning. Results are Feature Match Recall (FMR) in \%.}
\label{tab:fmr_lo_3dm}
\end{table}%

\subsection{3DMatch results details}

We present in Table~\ref{tab:supervised20} and Table~\ref{tab:supervised21} the details of results of 3DMatch in supervised learning for the different scenes of the dataset with $\tau_2 = 0.05$ and $\tau_2 = 0.2$.

\begin{table*}[ht]
\small
\centering
\begin{tabular}[t]{lccccccccc}
\toprule
Methods & Kitchen & Home 1 & Home 2 & Hotel 1 & Hotel 2 & Hotel 3 & Study & MIT Lab & Average \\
\midrule
3DSmoothNet~\cite{gojcic2018perfect} &97.0&95.5&89.4&96.5&93.3&98.2&94.5&93.5&94.7\\
FCGF~\cite{choy2019fully} &-&-&-&-&-&-&-&-&95.2\\
FCGF$^*$~\cite{choy2019fully} &99.0&99.4&91.8&98.2&97.1&98.1&96.6&100.0&97.5\\
D3Feat~\cite{bai2020d3feat} &-&-&-&-&-&-&-&-&95.8\\
MultiView~\cite{Li_2020_CVPR} &99.4&98.7&94.7&99.6& \bf{100} &100&95.5&92.2&97.5\\
SpinNet~\cite{ao2020SpinNet} &99.2&98.1& \bf{96.1} & 99.6 &97.1&100&95.6&94.8&97.6\\
DIP~\cite{Poiesi2021} &-&-&-&-&-&-&-&-& 94.8\\
GeDI~\cite{Poiesi2021gedi} &-&-&-&-&-&-&-&-& 97.9\\
MS-SVConv(1) &98.8&98.1&92.8&\bf{99.6}&99.0&100& \bf{96.6} &96.1&97.6\\
MS-SVConv(3) & \bf{99.6} & \bf{99.4} & 94.2 & 99.1 & 99.0 & \bf{100} & 95.9 & \bf{100} & \bf{98.4} \\
\bottomrule
\end{tabular}
\caption{Feature Match Recall with $\tau_2=0.05$ on 3DMatch in supervised learning. FCGF$^*$ means that we evaluate ourselves the original code with a symmetric test, before computing the FMR.}
\label{tab:supervised20}
\end{table*}%

\begin{table*}[ht]
\small
\centering
\begin{tabular}[t]{lccccccccc}
\toprule
Methods & Kitchen & Home 1 & Home 2 & Hotel 1 & Hotel 2 & Hotel 3 & Study & MIT Lab & Average \\
\midrule
3DSmoothNet~\cite{gojcic2018perfect} &62.8&76.9&66.3&78.8&72.1&88.9&72.3&64.9&72.9\\
FCGF~\cite{choy2019fully} &-&-&-&-&-&-&-&-&67.4\\
FCGF$^*$~\cite{choy2019fully} &91.1&91.7&78.4&94.2&90.4&90.7&85.3&76.6&87.3\\
D3Feat~\cite{bai2020d3feat} &-&-&-&-&-&-&-&-&75.8\\
MultiView~\cite{Li_2020_CVPR} &89.5&85.9&81.3&95.1& \bf{92.3} & \bf{94.4} &80.1&76.6&86.9\\
SpinNet~\cite{ao2020SpinNet} &-&-&-&-&-&-&-&-&85.7\\
DIP~\cite{Poiesi2021} &-&-&-&-&-&-&-&-& -\\
GeDI~\cite{Poiesi2021gedi} &-&-&-&-&-&-&-&-& -\\
MS-SVConv(1) &88.7&89.1&82.7&95.1&91.3&90.7&83.6&76.6&87.2\\
MS-SVConv(3) & \bf{95.8} & \bf{94.2} & \bf{83.7} & \bf{95.6} & 88.5 & 88.9 & \bf{87.0} & \bf{85.7} & \bf{89.9}\\
\bottomrule
\end{tabular}
\caption{Feature Match Recall with $\tau_2=0.2$ on 3DMatch in supervised learning. FCGF$^*$ means that we evaluate ourselves the original code with a symmetric test, before computing the FMR.}
\label{tab:supervised21}
\end{table*}%

\section{More experiments on the synergy between UDGE and MS-SVConv}
In this section, we present experiments on UDGE. We show that shared U-Net is better than unshared U-Net. In addition, we explain why we choose three U-Net for MS-SVConv. We show that three heads is a good trade-off between good generalization and computation. We also present an experiment to show that UDGE can be used for 100\% unsupervised training on the 3DMatch dataset. This results suggest that when the dataset is huge and the ground truth is not available, UDGE can be used to learn meaningfull features in an unsupervised fashion. We explain how we can choose the voxel size with UDGE and how we choose the size of crop. We show that for different voxel size, the results do not change a lot. However, for the size of crop, it is quite important and must be well chosen.
Finally, we show details results on the ETH dataset.

\subsection{To share or not to share?}
\begin{table}[ht]
\small
\centering
\tabcolsep=0.11cm
\begin{tabular}[t]{lccc}
\toprule
\multicolumn{4}{c}{ETH 8-scenes Dataset} \\
\midrule
Methods & FMR (\%) & SRE & Time (s)\\
\midrule
\multicolumn{4}{c}{Without UDGE} \\
MS-SVConv(1) & 56.4 & 151.0 & 0.24  \\
MS-SVConv(3) Unshared & 67.4 & 129.6 & 0.50 \\
MS-SVConv(3) & 76.8 & 82.2 & 0.52 \\
\midrule
\multicolumn{4}{c}{With UDGE} \\
MS-SVConv(1) & 87.5 & 44.0 & \bf{0.16} \\
MS-SVConv(3) Unshared & 93.5& 33.1 & 0.335\\
MS-SVConv(3) & \bf{93.6} & \bf{6.9} & 0.40 \\
\bottomrule
\end{tabular}
% S
\caption{Feature Match Recall (FMR) and median Scaled Registration Error (SRE) x1000 on ETH 8-scenes dataset with network pre-trained on 3DMatch. MS-SVConv(1) means MS-SVConv with 1 head and MS-SVConv(3) means MS-SVConv with 3 heads. Unshared means a multi-scale network with different weights at each scale. We only report the average time of descriptor extraction.}
\label{tab:unshared}
\end{table}%

We compare the results of the multi-scale by having shared weights or not between the U-Net. Table~\ref{tab:unshared} shows that when we do not share weights (i.e. three different U-Net for the three heads), we see that the generalization capacity results are much lower than the multi-scale with shared weights. Indeed, the gain is only +11\% with the unshared weights while it is +20.4\% on the Feature Match Recall for the shared weights during the transfer from the 3DMatch to ETH datasets. One reason for this may come from the fact that with unshared weights, each head specializes on a scale while the same network shared between scales learns to be robust to the notion of scale and will adapt better to a new dataset. As in multi-task learning, weights can be beneficial for every scale. We can notice that with UDGE, results are quite similar between shared and unshared networks (as it was the case for fine-tuning with two, three or four heads), but MS-SVConv with three shared heads still obtains better results. This experiment results explain why we choose shared U-net.

\subsection{Influence of the number of heads}

\begin{figure}
    \centering
    \includegraphics[width=0.45\textwidth]{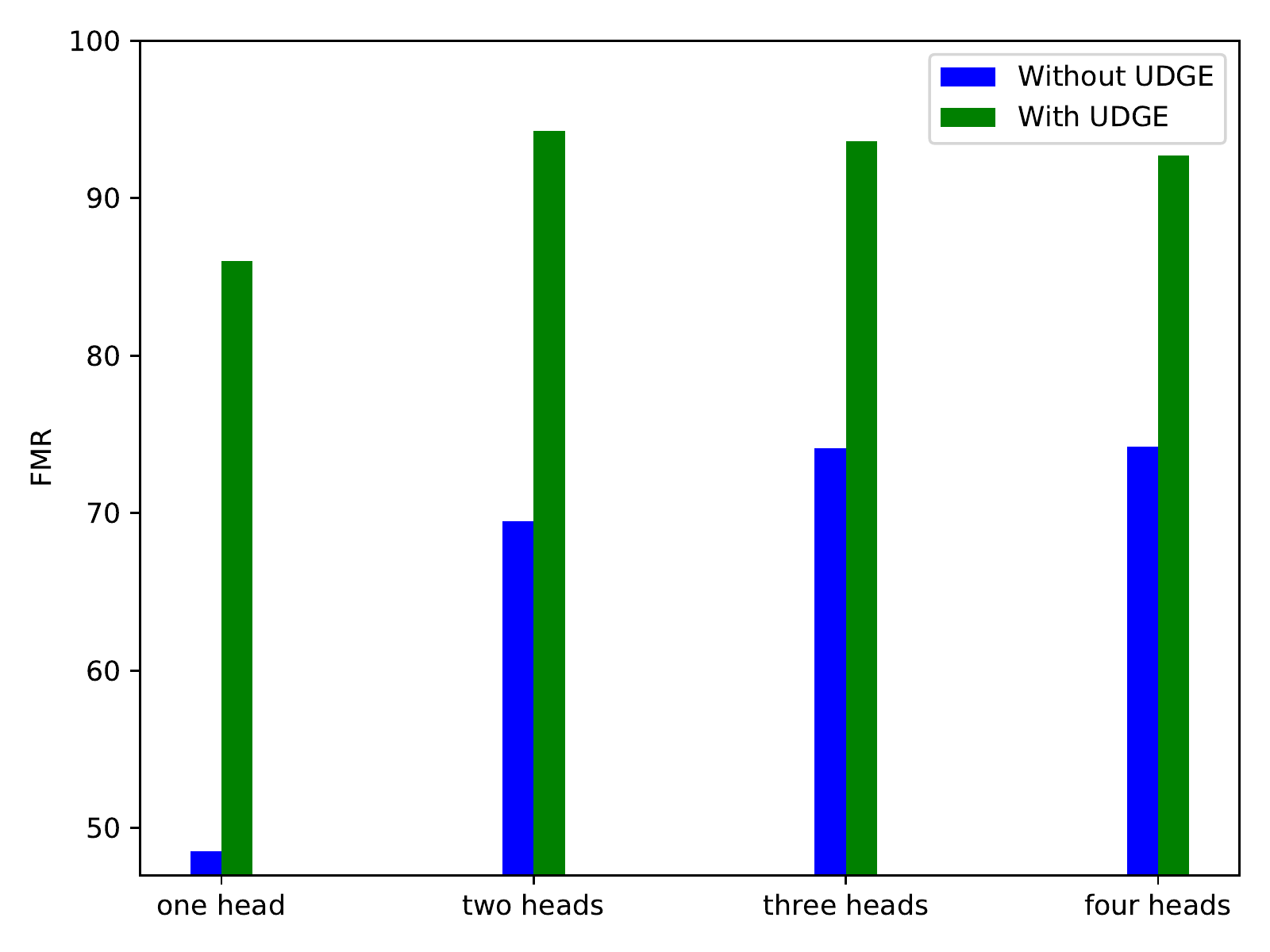}
    \caption{Feature Match Recall (FMR) on ETH 8-scenes dataset in function of the number of head without and with UDGE (model pre-trained on the ModelNet dataset).}
    \label{fig:influence_head}
\end{figure}

The different experiments on the the 3DMatch and the ETH dataset have shown the superiority of the proposed multi-scale network with three heads compared to one head in supervised training, for transfer after using UDGE. But how do we fix the number of heads? To answer this question, we train several networks on the ModelNet dataset with different number of heads and we compute the FMR on the ETH 8-scenes dataset without UDGE and with UDGE. As in previous experiments, the voxel size on the ETH dataset is set at 2 cm with no UDGE and 4 cm with UDGE (the length side of the voxels at the next scale is always 2 times that of the previous scale).
Figure~\ref{fig:influence_head} shows the FMR in function of the number of head. These results show that the more heads we add, the more the network is able to generalize (without UDGE) between the ModelNet, and the ETH dataset (results slightly decreases with 4 heads). With UDGE, we see that two heads is already enough to achieve excellent descriptor registration performance on ETH. Thus, three heads is a good trade-off to have a very good capacity for generalization on new datasets while having excellent performance with UDGE but keeping a sufficiently fast processing time (2.5 times slower between 3 heads and one head).

\subsection{100\% unsupervised training on 3DMatch with UDGE}
To show that our method can adapt to unseen scenes, and be applied in an unsupervised way, we realized an other experiment with the 3DMatch dataset. We first pre-train on the ModelNet dataset and then apply our unsupervised transfer learning strategy UDGE on the training set of the 3DMatch dataset. We finally evaluate MS-SVConv on the test set of the 3DMatch dataset.
Results in Table~\ref{tab:svsss} show that MS-SVConv trained in an unsupervised fashion is very close to supervised learning and comparable to other state-of-the-art methods, while being only trained on the synthetic dataset ModelNet. This is a second experiment which shows that the proposed method UDGE is not just over-fitting on the training set. 

\begin{table}[ht]
\small
\centering
\begin{tabular}[t]{lcc}
\toprule
Method & FMR ($\tau_2=0.05$) &  FMR ($\tau_2=0.2$)\\
\midrule
Supervised  & 98.4 & 89.9 \\
UDGE & 96.7 & 85.0 \\
\bottomrule
\end{tabular}
\caption{Comparison between supervised learning and unsupervised learning on 3DMatch. For unsupervised: model pre-trained on ModelNet and using UDGE on training set of 3DMatch (i.e. not using ground truth poses of 3DMatch). Model is MS-SVConv(3) for supervised and unsupervised results.}
\label{tab:svsss}
\end{table}%

\subsection{Influence of voxel side length for UDGE}

\begin{table}[ht]
\small
\centering
\begin{tabular}[t]{lccc}
\toprule
Voxel size (cm) & 2 & 4 & 6 \\
MS-SVConv(1) & 67.3 &87.5 & 91 \\
\midrule
Voxel size (cm) & 2, 4, 8 & 4, 8, 16 & 6, 12, 24 \\
MS-SVConv(3) & 90.9 & 93.6 & 93.8 \\
\bottomrule
\end{tabular}
\caption{Influence of the voxel size when applying our unsupervised UDGE method on ETH 8-scenes. Results are Feature Match Recall in \% with networks pre-trained on 3DMatch with voxel size = 2 cm for MS-SVConv(1) and voxel size = 2, 4, 8 cm for MS-SVConv(3).}
\label{tab:size_voxel}
\end{table}%

As presented in the article, networks based on sparse voxel convolution, such as MS-SVConv are sensitive to the choice of voxel size for discretization. We have already shown the interest of multi-scale in supervised training and for the transfer between datasets without UDGE (with a voxel size which must remain fixed). With UDGE, we can change the voxel size when we have a new dataset. For example, between 3DMatch and ETH, the scenes are much larger which pushes us to increase the size of the voxels to improve registration. We see this effectively with Table~\ref{tab:size_voxel} where we pre-trained our networks on the 3DMatch dataset and transfered on the ETH dataset.
With a single head, keeping the same voxel size as during pre-training, the network does not generate good decriptors. However, thanks to UDGE, by having a voxel size of 6 cm, the network with one head obtains much better results on ETH. Conversely, our network with three heads is much more robust to the variation in the size of the voxel and already obtains a very good score of 90.9\% FMR even without changing the size of the voxel between the 3DMatch, and the ETH  dataset (2, 4, 8 cm). 

\subsection{Influence of the size of the crop for UDGE}
\begin{figure}
    \centering
    \includegraphics[width=\linewidth]{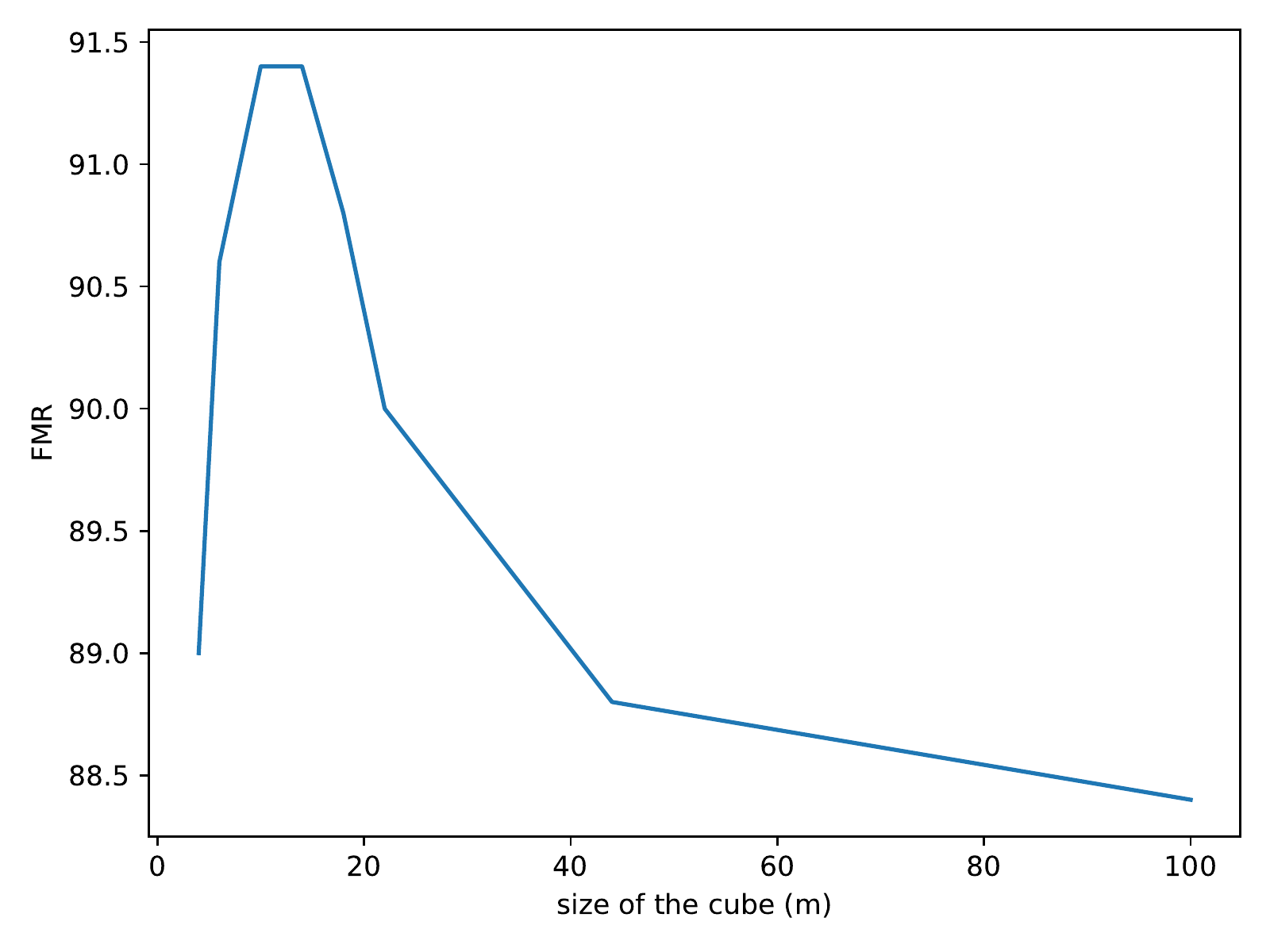}
    \caption{Feature Match Recall (FMR) of MS-SVConv (3) on ETH 8-scenes with respect to the size of the crop of our data generation for UDGE. Model pre-trained on ModelNet. These experiments have been done without periodic sampling}
    \label{fig:influence_crop}
\end{figure}

%Choosing the right size of the crop in UDGE is not obvious. 
We measured the influence of the size of the crop in our data generation for UDGE. Figure~\ref{fig:influence_crop} shows the FMR with respect to the size of crop on ETH 8-scenes dataset (model pre-trained on ModelNet). A big size of crop is equivalent to no crop at all. This experiment shows that this parameter is important for UDGE and must be chosen according to the dataset (see implementation details). 

%\begin{table}[ht]
%\centering
%\begin{tabular}[t]{lcccccccc}
%\toprule
%Size & 4m& 6m & 10m & 14m & 18m & 22m & 44m & 100m \\
%\midrule
%MS-SVConv(3) (old) & - & 92.3 & 92.6 & 91.9 & 91.6 & 91.4 & 90.2 & -  \\
%MS-SVConv(3)  & 89 & 90.6 & 91.4 & 91.4 & 90.8 & 90 & 88.8 & 88.4  \\
%\bottomrule
%\end{tabular}
%\caption{Feature Match Recall (FMR) with $\tau_2=0.05$ on ETH dataset, benchmark of Fontana~\etal~\cite{fontana2020benchmark} with respect to the size of patch for fine tuning we did not apply periodic sampling for this training}
%\label{tab:ethfontana}
%\end{table}%

\subsection{ETH results details}

We present in Table~\ref{tab:ethfontana} the details of results on the ETH 8-scenes dataset on the different scenes of the dataset. It shows that for the Hauptgebaude scene (scan of a university), results are much lower. Hauptegebaude is more challenging than others because it contains a lot of repetitive patterns
%. andit is a bigger scene with sparse element 
(see qualitative results in Figure~\ref{fig:qualitative_eth}).

\begin{table*}[ht]
\small
\centering
\begin{tabular}[t]{lccccccccc}
\toprule
Methods & Apart. & Gaz. Sum. & Gaz. Wint. & Haupt. & Plain & Stairs & Wood Aut. & Wood Sum. & Average \\
\midrule
\multicolumn{10}{c}{Classical methods} \\
FPFH~\cite{rusu_fast_2009} &2/33.1&0/29.2&0/23.6&0/320&0/64.6&4/38.5&0/95.1&0/76.6&0.75/85.1 \\
\midrule
\multicolumn{10}{c}{Patch based deep methods without UDGE}\\
MultiView~\cite{Li_2020_CVPR} & 83/14.8 & 42/9.4 & 52/8.3 & 47/262 & 17/21.4 & 65/15.2 & 13/12.1 & 22/9.8 &42.6/44.0\\
DIP~\cite{Poiesi2021} & 98/8.2 & 91/6.4 & 100/5.7 & \textbf{73}/5.7 & \textbf{96}/\textbf{9.3} & 97/\textbf{6.7} & 97/6.4 & 99/6.4 & \textbf{93.9}/\textbf{6.9} \\
\midrule
\multicolumn{10}{c}{U-Net based deep methods without UDGE} \\
D3Feat~\cite{bai2020d3feat} & 96/9.4 & 78/8.7 & 93/7.4 & 54/646 & 18/54.1 & 72/12.0 & 54/10.5 & 51/12.4 &64.5/95.0\\
MS-SVConv(1) & 96/8.8 & 65/8.6 & 79/7.4 & 31/554 & 14/579 & 63/26.5 & 51/10.1 & 52/10.5&56.4/151.0 \\
MS-SVConv(3) & 99/\textbf{8.1} & 85/7.7 & 98/5.6 & 43/533 & 36/76.1 & 84/10.5 & 81/8.7 & 88/8.0 & 76.8/82.2\\
\midrule
\multicolumn{10}{c}{U-Net based deep methods with UDGE} \\
MS-SVConv(1) & \textbf{100}/9.2 & 89/6.6 & 100/4.8 & 53/301 & 73/10.9 & 93/7.3 & 94/6.4 & 98/5.9 &87.5/44.0 \\
MS-SVConv(3) & 99/8.8 & \textbf{97}/\textbf{5.9} & \textbf{100}/\textbf{4.5} & 70/\textbf{5.5} & 85/9.8 & \textbf{98}/8.5 & \textbf{100}/\textbf{6.1} & \textbf{100}/\textbf{5.7} & 93.6/\textbf{6.9} \\
\bottomrule
\end{tabular}
\caption{Feature Match Recall (FMR) with $\tau_2=0.05$ and median Scaled Registration Error (SRE) x1000 per scene on ETH 8-scenes dataset, benchmark of Fontana~\etal~\cite{fontana2020benchmark}. All deep methods are pre-trained on 3DMatch. For every method, we compute the transformation using TEASER algorithm~\cite{yang2020teaser}. For FMR, higher is better and for SRE, lower is better.}
\label{tab:ethfontana}
\end{table*}%

\section{Implementation and Protocol}

\subsection{Implementation details}
To manage the experiments, we used the Pytorch Points 3D framework~\cite{tp3d}. This framework massively uses the hydra library to manage the hyperparameters, the architectures of the networks, and the data augmentations.
For the sparse convolution and sparse tensors, we use the implementation of Pytorch Points 3D that utilizes the torchsparse backend (implementated by Tang \etal~\cite{tang2020searching}). The implementation of Pytorch Points 3D also supports the MinkowskiEngine backend~\cite{choy20194d}, but sparse convolution in torchsparse is faster than in MinkowskiEngine. For all trainings, we use Stochastic Gradient Descent with momentum of 0.8 as optimizer and a batch size of 4. For pre-training, the number of epochs is 400 for ModelNet (around 48h), and 300 for the 3DMatch dataset (around 3 weeks), and the learning rate is $0.1$. For unsupervised transfer learning, the number of epochs is 200 for all datasets (around 2h for TUM, 18h for ETH, and 24h for 3DMatch), and the learning rate is $0.001$. 
%\footnote{more details here: \url{https://github.com/mit-han-lab/torchsparse}}.

On 3DMatch, ModelNet, and TUM, the size of the initial voxel is 2~cm and is doubled at every scale. For the ETH dataset (only when we use UDGE), the size of the initial voxel is 4~cm and also doubled at every scale (ETH dataset is sparser than 3DMatch). The choice of the voxel size depends on the point density.
The dimension of output descriptors is 32 for every dataset (as most previous published papers).

For the data generation of UDGE, the crop has a cubic shape, the center is a random point, and the size is 2~m for ModelNet, 3~m for 3DMatch and TUM and 10~m for ETH. %The size of the crop will depend on the size of the scene, and a good rule is to limit the crop to half of the size of the scene.
%to keep half of the size of scenes. 
As for the crop parameter, the periodic sampling parameters will change according to the dataset: for ETH, we uniformly sample $\alpha$ between 15\% and 30\% and the period $T$ between 4~cm and 16~cm. For the TUM dataset, we uniformly sample $\alpha$ between 10\% and 40\% and the period $T$ between 2~cm and 8~cm. For 3DMatch, we do not use periodic sampling as point clouds are uniform.

Finally, we perform classical data augmentation (same for all datasets) such as random rotation around all axes, random scale between 0.9 and 1.2 and random Gaussian noise with $\sigma$ equals to 0.7~cm for TUM and 3DMatch and 1~cm for ETH and ModelNet. 

To compute the transformation, we always use the TEASER algorithm~\cite{yang2020teaser} (with the main parameter noise bound fixed at $\tau_1=0.1$). We choose TEASER over RANSAC because, TEASER is faster. 
%But Every method has been tested in the same conditions. 
For every experiment, we sample randomly 5000 random points to evaluate point cloud matching.
The experiments were done on a computer with a GPU GTX 1080Ti and a CPU Intel(R) Core(TM) i7-4790K.

\subsection{Architecture of the network}
\begin{figure}
    \centering
    \includegraphics[width=\linewidth]{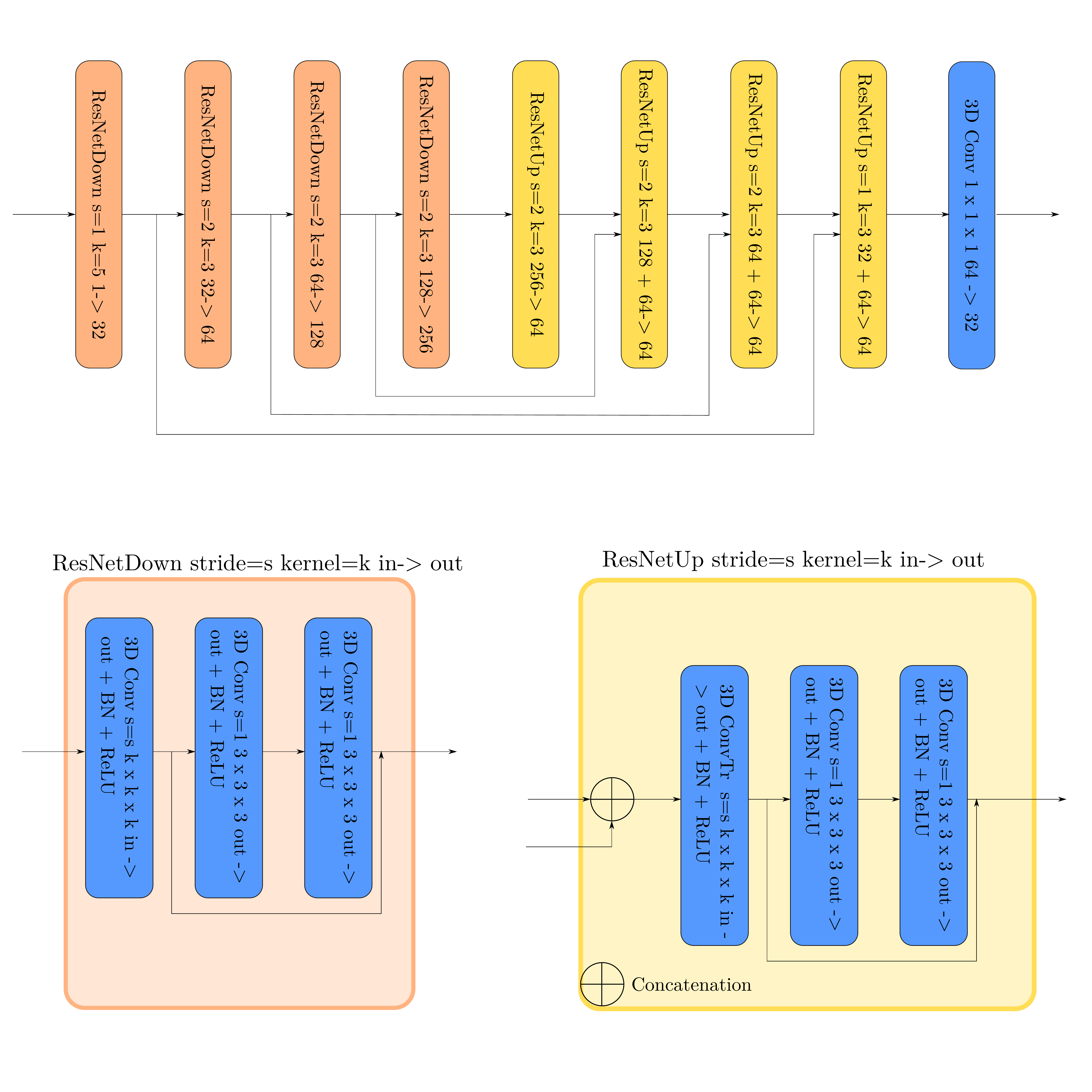}
    \caption{Architecture of one head of our U-Net network (BN means Batch Normalization). }
    \label{fig:architecture}
\end{figure}

In Figure~\ref{fig:architecture}, we report the detailed architecture of a U-Net of MS-SVConv. Because the weights are shared when using MS-SVConv with $S$ heads, the model still has around 9 million parameters in total. The only difference is in the additional MLP shared between all points with FC(S*d,d), which gives 3104 additional parameters for MS-SVConv(3), i.e. three heads with a descriptor of size 32. The architecture of MS-SVConv(1) and FCGF~\cite{choy2019fully} are almost similar but MS-SVConv has one supplementary ResBlock. MS-SVConv is also faster than FCGF: FCGF uses the library MinkowskiEngine while MS-SVConv uses the torchsparse library. For the evaluation, contrary to~\cite{bai2020d3feat, Poiesi2021, gojcic2018perfect, ao2020SpinNet}, FCGF~\cite{choy2019fully} does not perform a symmetric test when computing the FMR. This is why, results of FCGF are lower in term of FMR. 

\subsection{Protocol and metrics on ETH dataset}

\begin{table*}[ht]
\small
\centering
\begin{tabular}[t]{lccccc}
\toprule
Methods & Gaz. Sum. & Gaz. Wint & Wood Aut. & Wood Sum. & Average \\
\midrule
\multicolumn{6}{c}{Classical methods} \\
FPFH~\cite{rusu_fast_2009} & 38.6 & 14.2 & 14.8 & 20.8 & 22.1\\
SHOT~\cite{Salti2014SHOTUS} & 73.9 & 45.7 & 60.9 & 64.0 & 61.1\\
\midrule
\multicolumn{6}{c}{Patch based deep methods without UDGE} \\
3DSmoothNet~\cite{gojcic2018perfect} & 62.8 & 76.9 & 66.3 & 78.8 & 72.1\\
MultiView~\cite{Li_2020_CVPR} & 89.5 & 85.9 & 81.3 & 95.1 & 92.3\\
DIP~\cite{Poiesi2021} & 90.8 & 88.6 & 96.5 & 95.2 & 92.8\\
SpinNet~\cite{ao2020SpinNet} & 92.9 & 91.7 & 92.2 & 94.4 & 92.8\\
GeDI~\cite{Poiesi2021gedi} & \bf{98.9} & 96.5 & 97.4 &\bf{100} & 98.2 \\
\midrule
\multicolumn{6}{c}{U-Net deep methods without UDGE} \\
FCGF~\cite{choy2019fully} & 22.8 & 10.0 & 14.8 & 16.8 & 16.1 \\
D3Feat~\cite{bai2020d3feat} & 85.9 & 63.0 & 49.6 & 48.0 & 61.6\\
MS-SVConv(1) & 58.2 & 23.2 & 27.0 & 31.2 & 34.9\\
MS-SVConv(3) & 89.3 & 68.1 & 63.5 & 65.6 & 71.8\\
\midrule
\multicolumn{6}{c}{U-Net deep methods with UDGE} \\
MS-SVConv(1) & 86.4 & 90.0 & 85.2 & 90.4 & 88.0\\
MS-SVConv(3) & 95.7 & \bf{100} & \bf{100} & \bf{100} & \bf{98.9}\\
\bottomrule
\end{tabular}
\caption{Feature Match Recall (FMR) on ETH 4-scenes with the benchmark of Gojcic \etal~\cite{gojcic2018perfect} with $\tau_1 = 10$ cm and $\tau_2 = 0.05$. Results from other methods are from published papers. All networks are pre-trained on indoor 3DMatch.}
\label{tab:ethgojcic}
\end{table*}%

For our tests on the ETH dataset, we followed Fontana \etal's protocol~\cite{fontana2020benchmark} and Gojcic \etal's protocol~\cite{gojcic2018perfect}. Fontana \etal's protocol is more rigorous and uses eight scenes instead of four. Fontana \etal also introduce the Scaled Registration Error (SRE) to evaluate the transformation found by registration methods. Usually, we use a metric based on the rotation error or the translation error. Even if these metrics are useful, they suffer from several problems: there are two measures instead of one, so in some cases, comparison is not possible. Moreover, these metrics depends on whether the point cloud is centered or not. For example, if the point cloud is at 100~m from the center of rotation, a small rotation error of 2 degrees will bring a translation error of around 3~m.
To get the results of previous published methods on ETH 8-scenes (Fontana \etal's protocol~\cite{fontana2020benchmark}), we had to compute them by ourselves using online code. The following section will present the details of the choices made using the available codes of published methods.

\subsection{Experiment details for tested method on ETH-8 scenes}

We present the experimental details for the methods tested on ETH 8-scenes following the Fontana \etal's protocol~\cite{fontana2020benchmark}. We have varied the different parameters of all these methods to keep only the best results on ETH.

%\subsubsection{Iterative Closest Point~\cite{121791}}
%We use the implementation of Open3D~\cite{Zhou2018}. We down-sample the point cloud with a voxel size of 6 cm.

\subsubsection{FPFH~\cite{rusu_fast_2009}}
We use the implementation of Open3D~\cite{Zhou2018}. 
We down-sample the point cloud with a voxel size of 6 cm and we choose 5000 random points. To compute descriptors, we use a radius of 50~cm and use the 30 nearest neighbors to compute normals.

\subsubsection{DIP~\cite{Poiesi2021}}
For DIP, we first down-sample the point cloud with a voxel size of 6 cm. We use DIP on 5000 random points. The patch is a ball with a radius of $60 \sqrt{3} = 104$ cm. We use $p_\rho=0$. The model was provided by the authors (trained on 3DMatch).

\subsubsection{D3Feat~\cite{bai2020d3feat}}
For D3Feat, we down-sample the point cloud with a voxel size of 6 cm (as in the implementation). We use the Tensorflow implementation. We double the scale of the kernel points to increase the receptive field as done by Bai \etal~\cite{bai2020d3feat}. We use the model trained using the contrastive loss provided by the authors (we noticed that the model trained with the circle loss has worse results). 
We choose the 5000 points with the best score to compute the Feature Match Recall and the Scaled Registration Error. 

\subsubsection{MultiView~\cite{Li_2020_CVPR}}
For MultiView, we used the pre-trained model provided by the authors. We down-sample the point cloud with a voxel size of 6 cm and took 5000 random points.

\section{Qualitative results on the 3DMatch, ETH and TUM dataset}

\begin{figure*}[ht]
    \centering
    \includegraphics[width=\linewidth]{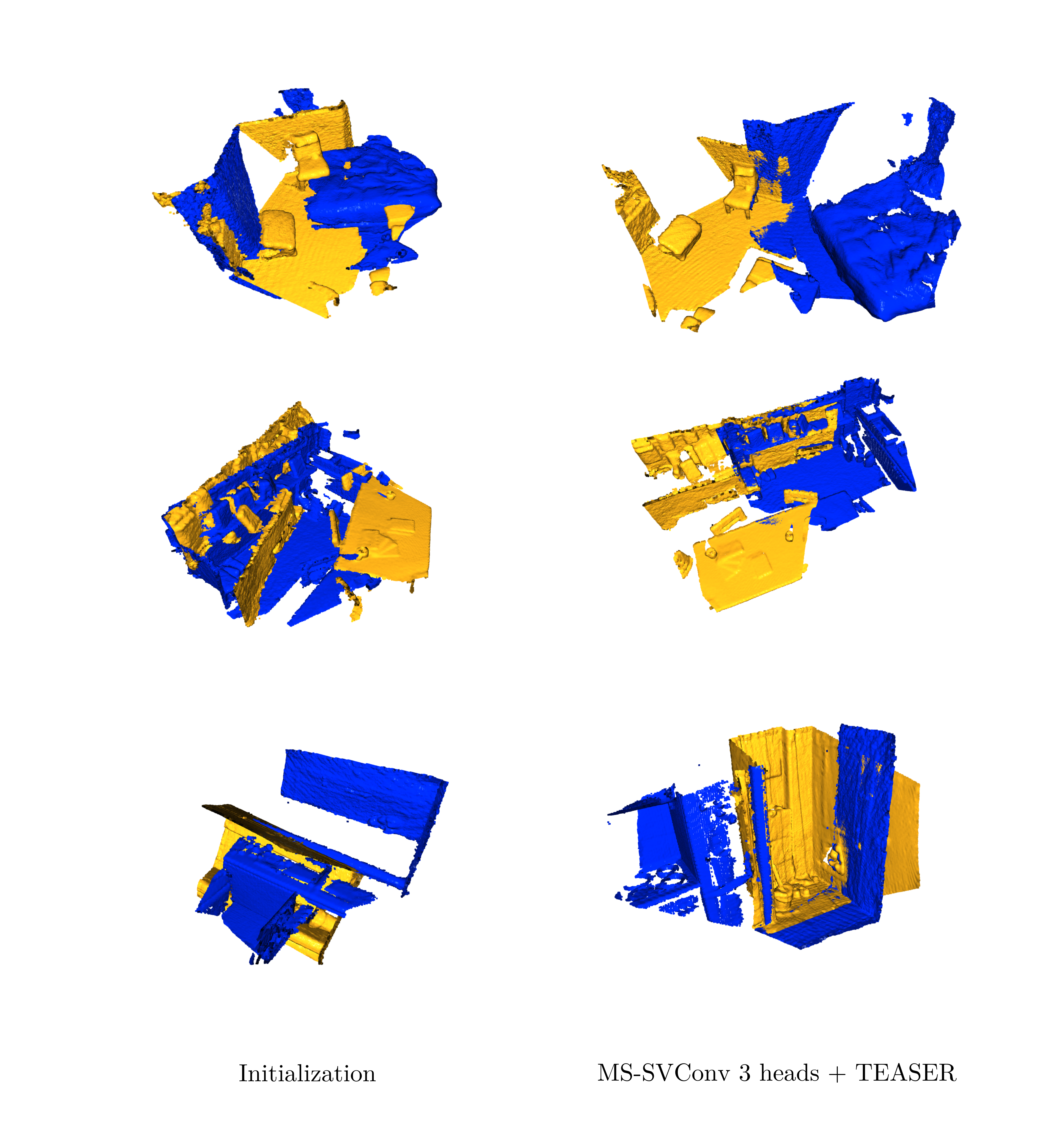}
    \caption{Qualitative results on 3DMatch (supervised learning). We can see that even with few structure and low overlap, MS-SVConv(3) can find the transformation between the pair of scenes.}
    \label{fig:qualitative_3dm}
\end{figure*}

\begin{figure*}[ht]
    \centering
    \includegraphics[width=\linewidth]{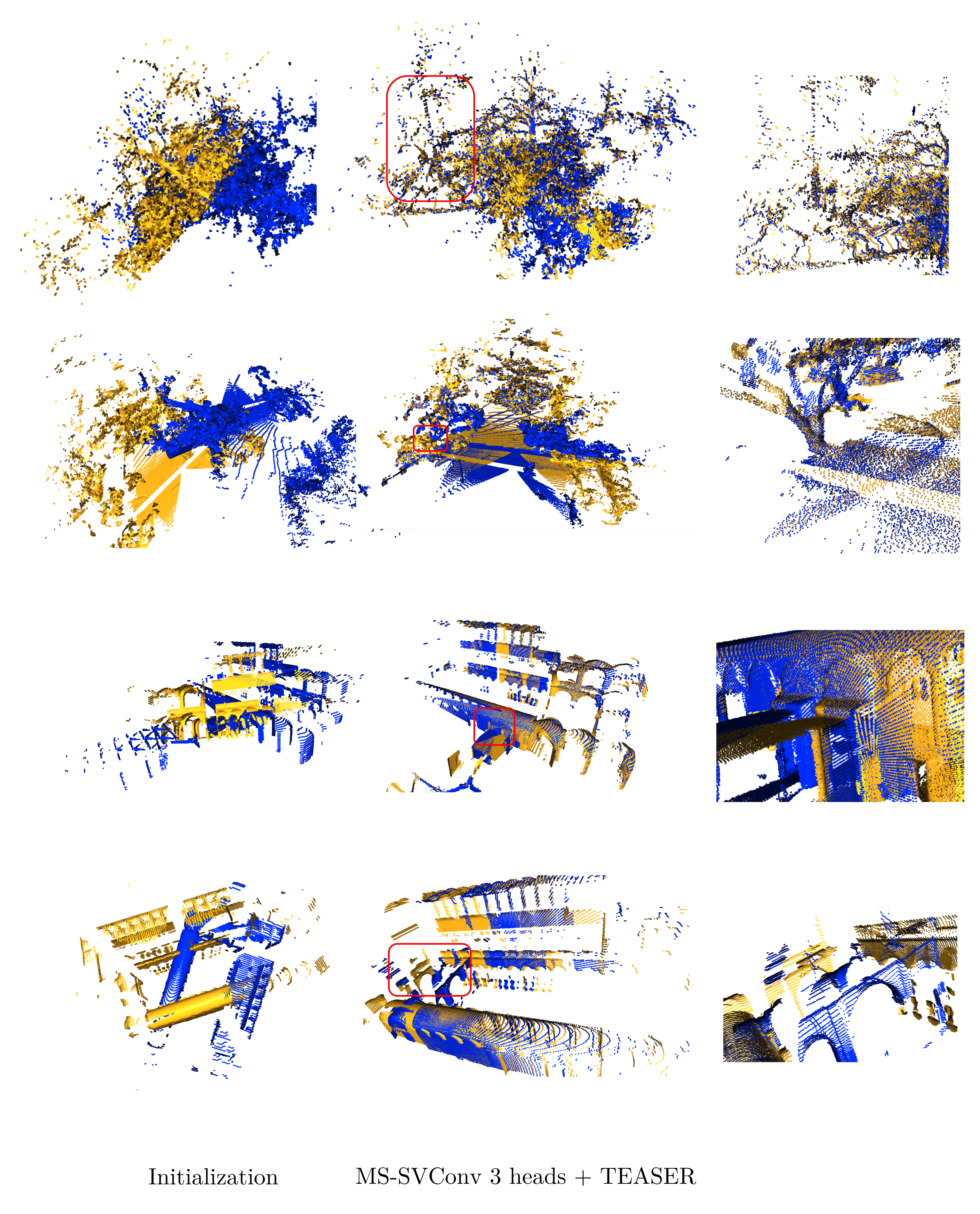}
    \caption{Qualitative results on ETH dataset (model pre-trained on 3DMatch and fine-tuned with UDGE on ETH). The last line shows a failure of MS-SVConv(3). }
    \label{fig:qualitative_eth}
\end{figure*}

\begin{figure*}[ht]
    \centering
    \includegraphics[width=\linewidth]{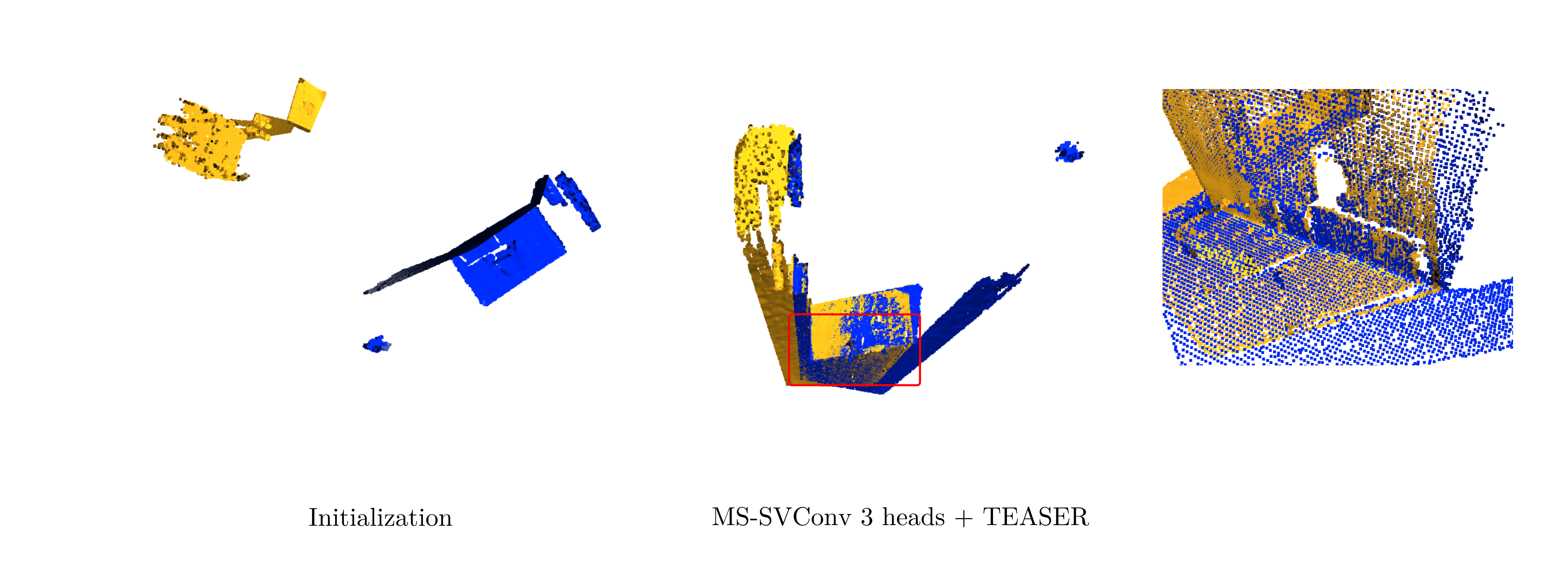}
    \caption{Qualitative results on TUM dataset (model pre-trained on 3DMatch and fine-tuned with UDGE on TUM).}
    \label{fig:qualitative_tum}
\end{figure*}

We show in Figures~\ref{fig:qualitative_3dm},~\ref{fig:qualitative_eth} and~\ref{fig:qualitative_tum} images of point cloud registration using MS-SVConv(3) for the 3DMatch, ETH, and TUM dataset.

\end{document}